\pdfoutput=1 

\documentclass[runningheads]{llncs}
 
\usepackage{eccv}

\usepackage{eccvabbrv}

\usepackage{graphicx}
\usepackage{booktabs}

\usepackage[accsupp]{axessibility}  %

\usepackage[pagebackref,breaklinks,colorlinks]{hyperref}

\usepackage{orcidlink}

\usepackage{caption}
\usepackage{amsmath}
\usepackage{amssymb}
\usepackage{booktabs}
\usepackage{makecell}
\usepackage[export]{adjustbox}
\usepackage[normalem]{ulem}
\usepackage{textpos}  %
\usepackage{wrapfig}
\usepackage{tabularx}
\usepackage{subcaption}
\usepackage{tabulary,multirow,overpic,xcolor,subfloat}
\usepackage{color, colortbl}

\usepackage{amsmath,amsfonts}
\def\bm#1{\mathbf{#1}}

\def\eqref#1{equation~\ref{#1}}

\def\1{\bm{1}}

\def\mC{{\bm{C}}}

\def\mI{{\bm{I}}}

\def\mM{{\bm{M}}}

\def\mT{{\bm{T}}}

\DeclareMathAlphabet{\mathsfit}{\encodingdefault}{\sfdefault}{m}{sl}
\SetMathAlphabet{\mathsfit}{bold}{\encodingdefault}{\sfdefault}{bx}{n}

\newcommand\citep[1]{\cite{#1}} %

\newcommand{\modelname}{SegGen\xspace}

\newcommand{\app}{\raise.17ex\hbox{$\scriptstyle\sim$}}

\newcolumntype{x}[1]{>{\centering\arraybackslash}p{#1pt}}
\newcolumntype{y}[1]{>{\raggedright\arraybackslash}p{#1pt}}
\newcommand{\dt}[1]{\fontsize{5pt}{0.1em}\selectfont (#1)}
\newlength\savewidth\newcommand\shline{\noalign{\global\savewidth\arrayrulewidth
  \global\arrayrulewidth 1pt}\hline\noalign{\global\arrayrulewidth\savewidth}}
\newcommand{\tablestyle}[2]{\setlength{\tabcolsep}{#1}\renewcommand{\arraystretch}{#2}\centering\footnotesize}
\makeatletter\renewcommand\paragraph{\@startsection{paragraph}{4}{\z@}
  {.5em \@plus1ex \@minus.2ex}{-.5em}{\normalfont\normalsize\bfseries}}\makeatother

\DeclareMathAlphabet\mathbfcal{OMS}{cmsy}{b}{n}

\definecolor{Gray}{gray}{0.5}
\definecolor{LightCyan}{HTML}{dae9f4}

\begin{document}

\title{SegGen: Supercharging Segmentation Models with Text2Mask and Mask2Img Synthesis} 

\titlerunning{SegGen}

\authorrunning{Hanrong Ye et al.}
\author{%
 Hanrong Ye$^1$, Jason Kuen$^2$, Qing Liu$^2$, Zhe Lin$^2$, Brian Price$^2$, Dan Xu$^{1}$\\
 }

\institute{CSE, HKUST \and Adobe Research \\
}

\maketitle

\begin{abstract}
We present \modelname, a new data generation approach that pushes the performance boundaries of state-of-the-art image segmentation models.
One major bottleneck of previous data synthesis methods for segmentation is the design of ``segmentation labeler module'', which is used to synthesize segmentation masks for images~\cite{wu2023diffumask}. The segmentation labeler modules, which are segmentation models by themselves, bound the performance of downstream segmentation models trained on the synthetic masks. These methods encounter a  ``chicken or egg dilemma'' and thus fail to outperform existing segmentation models. 
To address this issue, we propose a novel method that \textbf{\textit{reverses}} the traditional data generation process: we first (i) generate highly diverse segmentation masks that match real-world distribution from text prompts, and then (ii) synthesize realistic images conditioned on the segmentation masks. In this way, we avoid the need for any segmentation labeler module. 
\modelname integrates two data generation strategies, namely MaskSyn and ImgSyn, to largely improve data diversity in synthetic masks and images. 
Notably, the high quality of our synthetic data enables our method to outperform the previous data synthesis method~\cite{wu2023diffumask} by +25.2 mIoU on ADE20K when trained with pure synthetic data. 
On the highly competitive ADE20K and COCO benchmarks, 
our data generation method markedly improves the performance of state-of-the-art segmentation models in semantic segmentation, panoptic segmentation, and instance segmentation.
Moreover, experiments show that training with our synthetic data makes the segmentation models more robust towards unseen data domains, including real-world and AI-generated images.
\end{abstract}

\begin{figure*}[h]
    \centering
\includegraphics[width=1\linewidth]{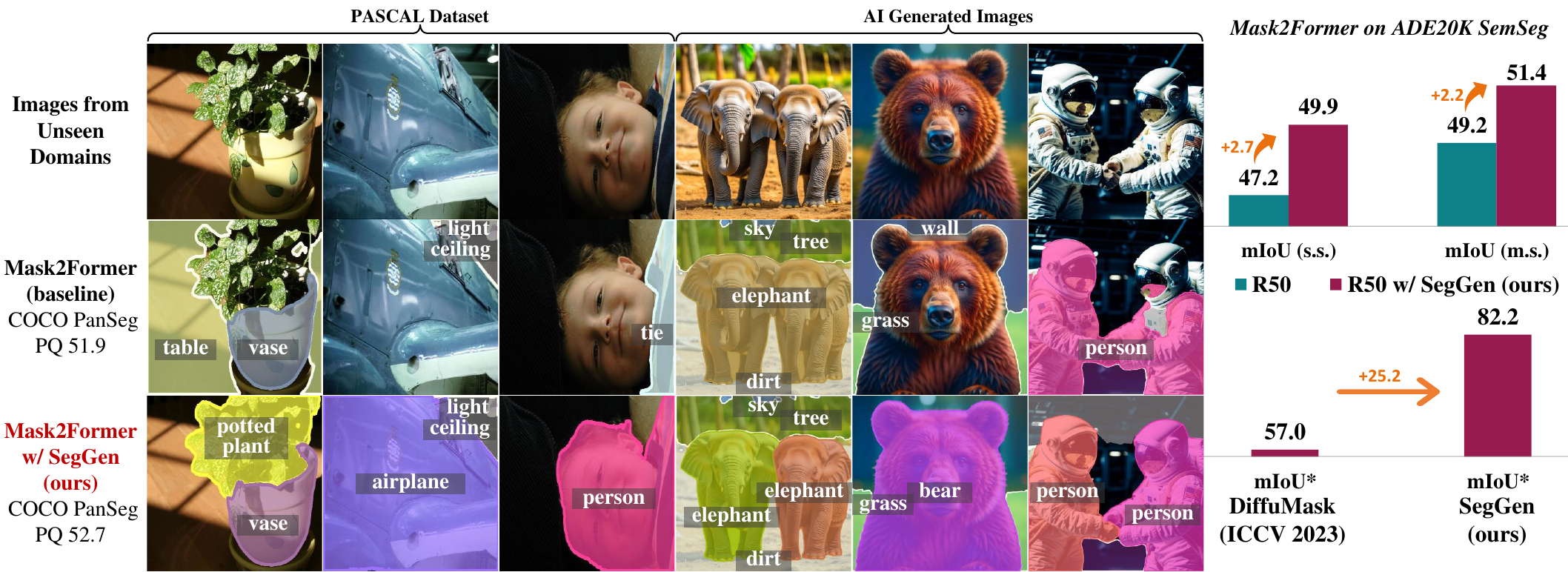}
\vspace{-20pt}
\caption{\textbf{Effectiveness of \modelname on various data domains:}
Through training with synthetic data generated by the proposed \modelname, we significantly boost the performance of state-of-the-art segmentation model Mask2Former~\citep{cheng2021mask2former} on evaluation benchmarks including ADE20K~\citep{ade20k_sceneparse_150} and COCO~\citep{lin2014coco}, whilst making it more robust towards challenging images from other domains~(the three columns on the left are from PASCAL~\citep{everingham2015pascal}; the three on the right are synthesized by image generation model Kandinsky 2~\citep{kandinsky}).
SegGen outperforms the previous best data generation method~(DiffuMask~\cite{wu2023diffumask}) by a huge margin when models are trained on pure synthetic data. ``mIoU*'' is the average IoU metric defined by~\cite{wu2023diffumask} which focuses on three common classes.
}
\label{fig:teaser}
\vspace{-20pt}
\end{figure*}

\section{Introduction}
Image segmentation explores the identification of objects in visual inputs at the pixel level. Based on the different emphases on category and instance membership information, researchers have divided image segmentation into several tasks~\citep{long2015fully,chen2014semantic,kirillov2017panoptic,entityseg}. 
For example, semantic segmentation studies pixel-level understanding of object categories, instance segmentation focuses on instance grouping of pixels, while panoptic segmentation considers both.
For all these segmentation tasks, obtaining high-quality annotation is challenging as every individual pixel requires human labeling, and a single image can contain millions of pixels.
Therefore, compared to other public datasets like ImageNet-21K~(with around 14M images), the prevailing human-annotated segmentation datasets are notably smaller. For example, ADE20K dataset~\citep{zhou2017ade20k} contains about 20K images in its training split, while COCO~\citep{lin2014coco} has around 118K training images. 
Although there has been significant development in the structure of segmentation models~\citep{zhao2017pspnet,li19,zhang2021knet,cheng2021maskformer,taskprompter2023}, the limited size of training data hinders further performance enhancements and results in inadequate generalization ability to handle images from unfamiliar domains, such as those from other scenes or synthesized by generative models as shown in the second row of Fig.~\ref{fig:teaser}.

Inspired by the recent success of image generation~\citep{karras2019style,dhariwal2021diffusion}, 
researchers start to explore using generative models to enhance image segmentation~\citep{baranchuk2021label}.
A representative direction of this research focuses on synthesizing segmentation training data in a cost-effective manner~\citep{li2022bigdatasetgan}.
In related methods, the synthetic segmentation masks are obtained from some manually designed ``segmentation labeler modules''. The segmentation labeler modules are essentially segmentation models by themselves. They are either small-scale segmentation networks as in DatasetGAN~\citep{datasetgan} and Grounded Diffusion~\citep{li2023grounded}, or post-processing methods on the image features as in DiffuMask~\citep{wu2023diffumask}.
The performance of the downstream segmentation models is constrained by the quality of the synthetic masks they are trained on, which in turn relies on the capabilities of those segmentation labeler modules. 
This is a ``chicken or egg dilemma'' and the performance bottleneck is the segmentation labeler module.
Therefore, while they achieve encouraging success in settings with limited training data, their methods fail to deliver notable enhancements for the state-of-the-art models on the most popular segmentation benchmarks in settings using the complete training sets.

\begin{figure*}[!t]
    \centering
\includegraphics[width=1\linewidth]{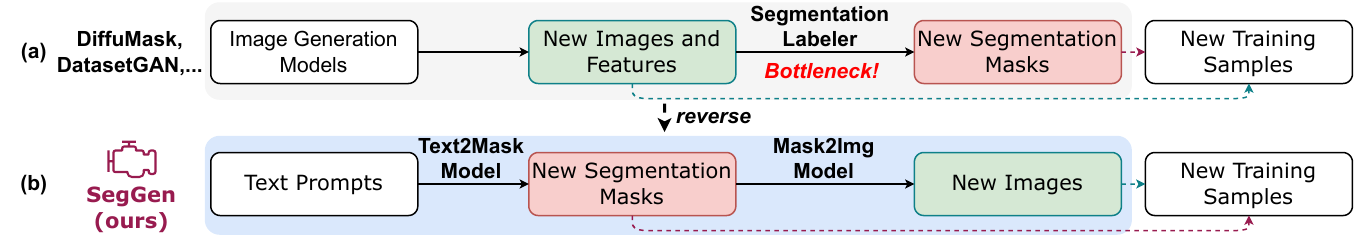}
\vspace{-20pt}
\caption{\textbf{Comparison with previous data generation methods for segmentation:}
\textbf{(a)} Earlier methods~\cite{wu2023diffumask, datasetgan} rely on segmentation labeler modules to produce segmentation masks for synthetic images. However, the performance of these downstream segmentation models, trained on synthetic data, is bounded by the capacity of the segmentation labeler modules. The segmentation labeler is a major bottleneck for the quality of the generated data.
\textbf{(b)} We design a reverse pipeline: we first create diverse new masks from text prompts via a proposed Text2Mask generation model and then synthesize images conditioned on the segmentation masks. This methodology avoids any usage of segmentation labeler networks, resulting in significantly improved quality of our synthesis data.
}
\label{fig:motivation}
\vspace{-15pt}
\end{figure*}

On the other hand, a handful of works~\citep{zhang2023adding,zhao2023uni} have demonstrated that high-quality synthetic images can be generated by %
conditioning text-to-image models on %
dense input maps (\textit{e.g.,} segmentation masks or canny edge). Moreover, these dense-input conditional models provide a strong alignment between the dense input label maps and generated images. These two qualities motivate us to leverage the powerful capabilities of such models for effective segmentation data generation. %
To this end, we propose a novel segmentation data generation method, coined as \textbf{\modelname}, for generating high-quality segmentation training data.
We reverse the traditional data generation pipeline for segmentation: we first synthesize high-quality segmentation masks from a proposed text-to-mask generation model, and then generate images based on the segmentation masks using a mask-to-image generation model.
In this new framework, we avoid using any segmentation labeler modules, which results in significantly improved data quality. Fig.~\ref{fig:motivation} shows the difference between our method and previous data generation methods for segmentation~\cite{wu2023diffumask,datasetgan}.

As each training sample of image segmentation consists of two components: %
segmentation masks and the corresponding image, we develop two novel data generation approaches, emphasizing improvements in two distinct aspects of data diversity: (i) segmentation masks and (ii) images.
The first data generation approach, named \textbf{MaskSyn}, centers around the generation of new segmentation masks. It learns a text-to-mask~(Text2Mask) generation model to produce completely new segmentation masks given text prompts. Then, it learns a mask-to-image~(Mask2Img) generation model to synthesize images that align with the synthetic segmentation masks.
The second data generation approach, named \textbf{ImgSyn}, utilizes the above-mentioned Mask2Img model to synthesize new images given human-annotated segmentation masks. 
With MaskSyn and ImgSyn, we can readily generate a vast array of diverse and high-quality synthetic training data.
The combined synthetic data is used to train segmentation models in conjunction with real training samples from human-annotated datasets.
Experiments show that our synthetic data can significantly boost the performance of the image segmentation models on challenging benchmarks including ADE20K semantic segmentation, COCO panoptic segmentation, and COCO instance segmentation, achieving new state-of-the-art performances without using extra human-annotated data. 
Notably, SegGen remarkably outperforms the previous segmentation data generation method~\cite{wu2023diffumask} by \textbf{+25.2} mIoU when trained with pure synthetic data on ADE20K. 
Our method boosts the mIoU of Mask2Former by +2.7 (R50) and +1.3 (Swin-L) on the ADE20K semantic segmentation benchmark.
Furthermore, segmentation models trained on our synthetic data exhibit a remarkably stronger ability to generalize across unfamiliar image domains, including real images from other distributions and machine-generated images.

We summarize the contribution of this work in three points:
    \textbf{(i)} We propose a revolutionary generation framework that reverses the traditional segmentation data generation pipeline and solves the ``chicken or egg dilemma''. The new framework enables us to produce high-quality segmentation training data at scale, thus enabling the training of more powerful image segmentation models.
    \textbf{(ii)}  We introduce two effective generative models, one for text-to-mask generation and the other for mask-to-image generation. Based on these models, we propose two novel segmentation training data generation approaches, namely MaskSyn and ImgSyn. They significantly improve the data diversity, with MaskSyn focusing on new segmentation masks and ImgSyn on new images.
    \textbf{(iii)} \modelname successfully improves the performance of the leading-edge segmentation models across the highly competitive benchmarks on ADE20K and COCO. 
    Moreover, \modelname enhances the generalization ability of segmentation models towards unseen image domains.
    Rigorous experiments, including ablation study and peer comparison, strongly suggest the effectiveness of the proposed method.

\section{Related Work}

\noindent\textbf{Generation for Segmentation}
Image segmentation is one of the most studied visual perception problems~\citep{li2023transformersurvey}.
In recent years, there has been a surge in efforts to harness the capabilities of generative models for segmentation tasks. These efforts can be broadly classified into three categories based on their methodologies:
(i) Extracting visual features from generative models for segmentation~\citep{baranchuk2021label,xu2023open,PNVR_2023_ICCV,zhao2023unleashing}.
These methods harvest the strong representation ability of diffusion models trained on large-scale datasets but are limited in the precision of predicted masks.
(ii) Formulating segmentation tasks directly as generative models~\citep{chen2022generalist,ji2023ddp,wang2023dformer}.
Although these methods propose exciting new model architectures, they usually consume higher computational costs while showing unimproved performance compared with conventional segmentation models.
(iii) Synthesizing segmentation training data using generative models~\citep{datasetgan,li2022bigdatasetgan,wu2023diffumask,li2023grounded,wu2023datasetdm,xie2023mosaicfusion,datasetdiffusion2023}. 
While these methods have showcased commendable results against their respective baselines especially when the training data is highly limited, they have yet to exhibit notably superior performance on the most rigorous benchmarks including ADE20K (150 categories)~\citep{zhou2017ade20k} and COCO (133 categories)~\citep{lin2014coco} under fully-supervised setting. 
A primary concern with these techniques is the subpar quality of the generated segmentation masks. This stems from their dependence on the bottleneck segmentation labeler modules during the synthetic mask generation process. 
Therefore, a concurrent work~\cite{freemask} abandons the generation of segmentation masks and directly generates images based on existing masks, which is highly limited in mask diversity.
In contrast, our data generation workflow reverses the traditional data generation pipeline: we first generate segmentation masks from text, and then synthesize images based on segmentation masks. Our method avoids the need for any segmentation labeler which results in largely improved data quality.
Our SegGen can strongly enhance the existing state-of-the-art segmentation models on different challenging benchmarks.

\noindent\textbf{Conditional Image Synthesis}
Within the realm of conditional image synthesis, Generative Adversarial Networks~\citep{goodfellow2020generative,brock2018large,karras2019style,kang2023scaling,Sauer2023ICML}, Variational Autoencoders~\citep{kingma2013auto}, and Diffusion Models~\citep{sohl2015deep,dhariwal2021diffusion,saharia2022photorealistic,ho2020denoising} have been at the forefront. Recently, the open-source research community has shown a burgeoning interest in the latent diffusion-based Stable Diffusion (SD) series for text-to-image synthesis~\citep{rombach2022ldm}. The most recent iteration, SDXL, introduced by~\cite{podell2023sdxl}, expands the model capacity, yielding significantly enhanced results. Therefore, we build SegGen upon SDXL.
Regarding mask-to-image generation models~\citep{lhhuang2023composer,zeng2023scenecomposer,li2023gligen}, ControlNet~\citep{zhang2023adding} and T2I-Adapter~\citep{mou2023t2i} suggest freezing the parameters of the SD model and introducing a set of more compact, learnable modules. This approach enables image generation conditioned on the given segmentation masks. 
We adopt the structure of ControlNet in our mask-to-image generation model.
\cite{zheng2023changen} propose a method for synthesizing novel images based on remote sensing change events.
To the best of our knowledge, there have been no prior endeavors on text-to-mask generation.

\section{Method}
\subsection{Overview}
\modelname is designed to synthesize high-quality training samples for improving the performance of segmentation models.
The overall workflow is shown on the left of Fig.~\ref{fig:framework}. We first train \modelname with human-annotated training samples from public datasets. 
After training, we use \modelname to produce new segmentation training samples at scale. The generated training samples are incorporated into the training process of segmentation models to enhance the model performance. %

\subsection{Models in \modelname}
\begin{figure*}[!t]
    \centering
\includegraphics[width=1\linewidth]{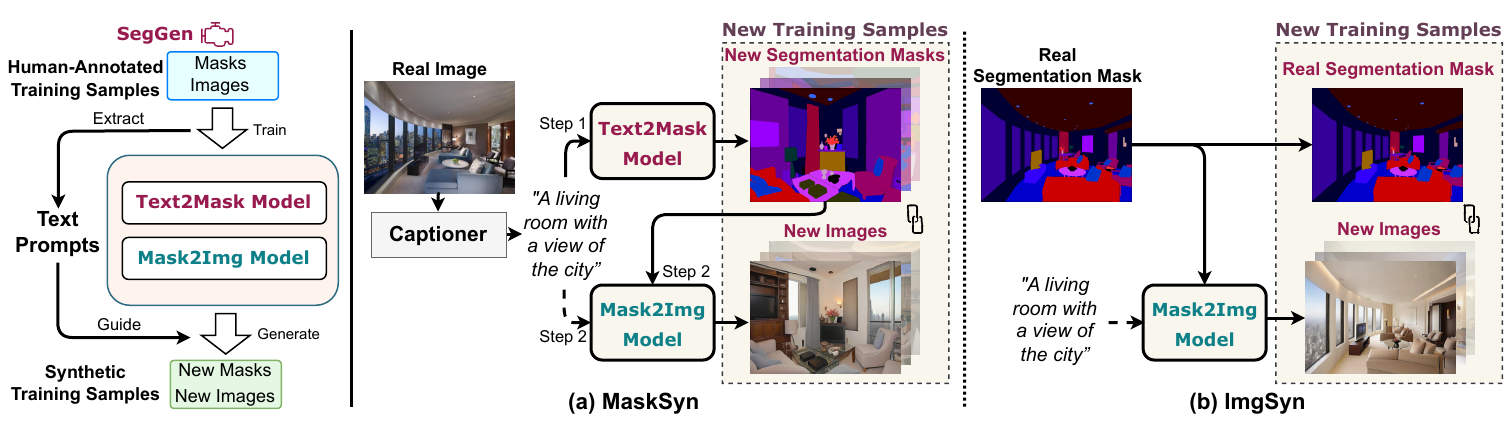}
\vspace{-20pt}
\caption{\textbf{Illustration of the workflow of our proposed \modelname:} We introduce two generative models: a text-to-mask~(Text2Mask) generation model and a mask-to-image~(Mask2Img) generation model, based on which we design two approaches for synthesizing segmentation training samples: MaskSyn and ImgSyn.
\textbf{(a)}~MaskSyn focuses on generating new segmentation masks. It first extracts the caption of the real image as a text prompt and uses it to generate new masks with the Text2Mask model. Then, the new masks and text prompt are fed into the Mask2Img model to produce the corresponding new images.
\textbf{(b)}~ImgSyn focuses on the synthesis of new images. It directly inputs human-labeled masks and text prompts into the Mask2Img model to generate new images.
}
\label{fig:framework}
\vspace{-15pt}
\end{figure*}

As shown in Fig.~\ref{fig:framework}, we first utilize a captioner model to extract captions of the real training images as text prompts from the target dataset. The text prompts will condition the data generation process.
Then, two conditional generative models are introduced: a text-to-mask~(Text2Mask) generation model and a mask-to-image~(Mask2Img) generation model. 
Both generative models are built upon the SDXL model~\citep{podell2023sdxl} which provides top-notch image generation quality. 

\noindent\textbf{Captioner Model}
To obtain image captions of existing training samples, we employ BLIP2-FlanT5$_\textrm{xxl}$~\citep{li2023blip2} model, which is a state-of-the-art vision-language model. We feed the text prompt \texttt{``Question: What are shown in the photo? Answer:''} and image as input to the model.
The responses serve as text prompts to condition the following generation process.

\noindent\textbf{Text2Mask Model} 
We design a Text2Mask model in \modelname for generating diverse segmentation masks based on given text prompts.
To leverage the generation capacity of text-to-image generation models pre-trained on large-scale datasets, we encode the segmentation masks~(the pixel values are category IDs) as three-channel RGB-like color maps, where one color represents a certain category. 
From our experiments, the color map reconstructed by VAE~\citep{rombach2022ldm} in SDXL appears almost indistinguishable from the original input as shown in the supplementary materials. %
Therefore, we can directly fine-tune the text-to-image SDXL-base model with [text, segmentation color map] training pairs, which are from the public image segmentation dataset~(\textit{e.g.} ADE20K).
During sampling, our Text2Mask model can generate diverse color maps conditioned on text prompts, which are subsequently converted into segmentation masks.
Formally, suppose the input text prompt is $\mT$, the target height and width are $H$ and $W$, the synthesized color map is $\mC_{\textrm{syn}}\in \mathbb{W}^{H\times W \times 3}$, and the synthesized segmentation map~(with $N$ masks) is $\mM_{\textrm{syn}}\in \mathbb{W}^{H\times W\times N}$, the generation process is:
\begin{equation}
\begin{aligned}
    \mC_{\textrm{syn}} &= \textrm{Text2Mask} (\mT), \\
    \mM_{\textrm{syn}} &= f_{\textrm{color}\rightarrow \textrm{mask}} (\mC_{\textrm{syn}}),
\end{aligned}
\label{eq:txt2mask}
\end{equation}
where $f_{\textrm{color}\rightarrow \textrm{mask}}: \mathbb{W}^{H\times W \times 3} \rightarrow \mathbb{W}^{H\times W \times N}$ is the function that projects the color maps to segmentation masks. %
Specifically, for each pixel on the color maps, we identify its nearest color (in Euclidean space) in the aforementioned lookup table, and assign the corresponding class to the pixel in the segmentation masks.

\noindent\textbf{Mask2Img Model}
The goal of the Mask2Img model is to synthesize new images that %
align well with the given segmentation masks and text prompts. Specifically we adopt the structure of  ControlNet~\citep{zhang2023adding}: we freeze the pre-trained weights of the SDXL-base model and train an additional side network for mask-conditioned image generation. %
It simultaneously keeps the generalization ability of the pre-trained diffusion model and provides excellent controllable generation ability. 
The Mask2Img model is trained with the [text, segmentation color map, image] triplets gathered from the training splits of the target datasets.
We denote the input segmentation map as $\mM \in \mathbb{W}^{H\times W \times N}$, the color map as $\mC \in \mathbb{W}^{H\times W \times 3}$, the synthetic image as $\mI_{\textrm{syn}}\in \mathbb{W}^{H\times W \times 3}$, and the generation process is:
\begin{equation}
\begin{aligned}
    \mC & = f_{\textrm{mask}\rightarrow \textrm{color}} (\mM), \\
    \mI_{\textrm{syn}} &= \textrm{Mask2Img} (\mT, \mC),
\end{aligned}
\end{equation}
where $f_{\textrm{mask}\rightarrow \textrm{color}}: \mathbb{W}^{H\times W \times N} \rightarrow \mathbb{W}^{H\times W \times 3}$ is the function to convert the segmentation masks into a color map.
For semantic segmentation, the value of each pixel on the segmentation mask corresponds to a category ID, allowing us to convert the masks directly into an RGB color map using a pre-defined lookup table.
For panoptic and instance segmentation, after mapping the category IDs to color maps, it is essential to outline each segment with a special edge color on the color map. This ensures the model recognizes the specific instance it belongs to.
The segmentation map $\mM$ can be human-annotated or synthetic~(\textit{i.e.}, $\mM_{\textrm{syn}}$ from Text2Mask Model as shown in Eq.~\ref{eq:txt2mask}).

\begin{figure*}[!t]
    \centering
     \includegraphics[width=1.\linewidth]{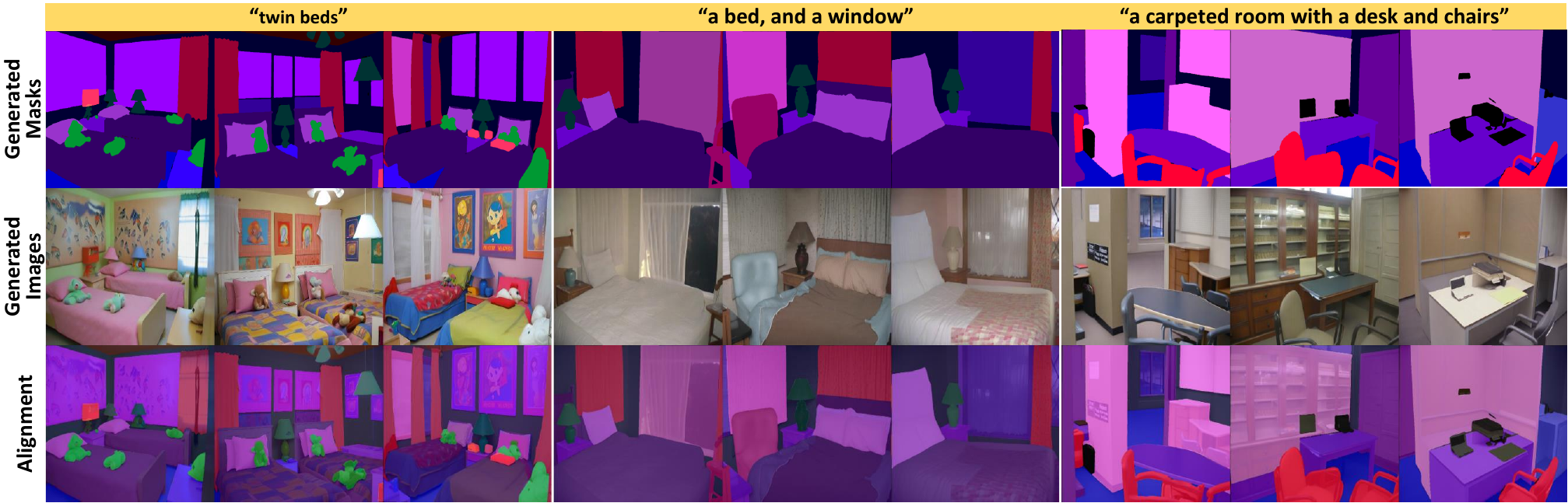}
     \vspace{-20pt}
     \caption{\textbf{Generated samples by MaskSyn on ADE20K:} The third row overlays the mask and the image together to demonstrate the alignment between them. The generated segmentation masks and images demonstrate high perceptual quality and excellent alignment (see more samples in supplementary materials).}
     \label{fig:vis_masksyn}
\vspace{-5pt}
\end{figure*}

\begin{figure*}[!t]
    \centering
     \includegraphics[width=1.\linewidth]{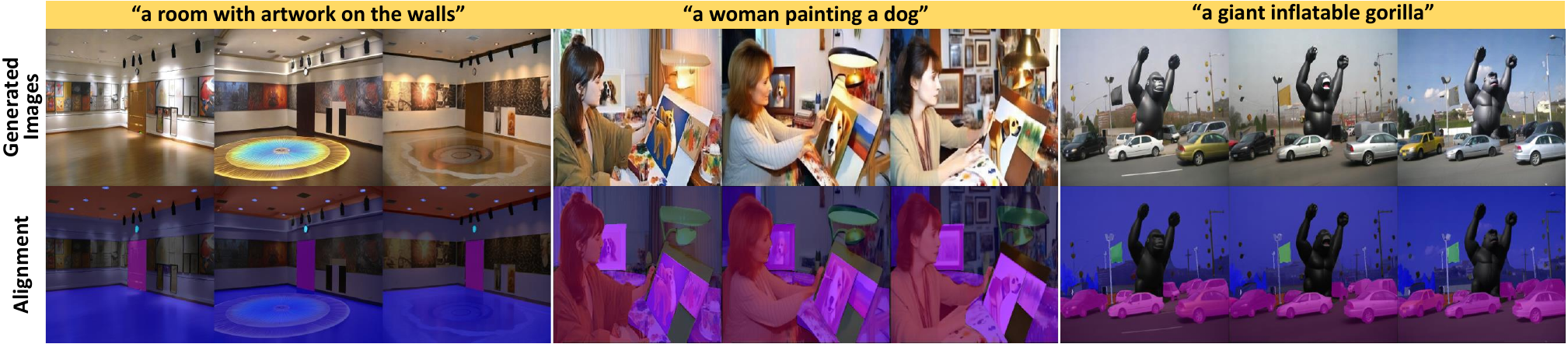}
     \vspace{-20pt}
     \caption{\textbf{Generated samples by ImgSyn on ADE20K:} The generated images exhibit remarkable realism and align well with the human-labeled masks and text prompts~(see more samples in supplementary materials).}
     \label{fig:vis_imgsyn}
\vspace{-10pt}
\end{figure*}

\begin{figure*}[!t]
    \centering
    \includegraphics[width=1.\linewidth]{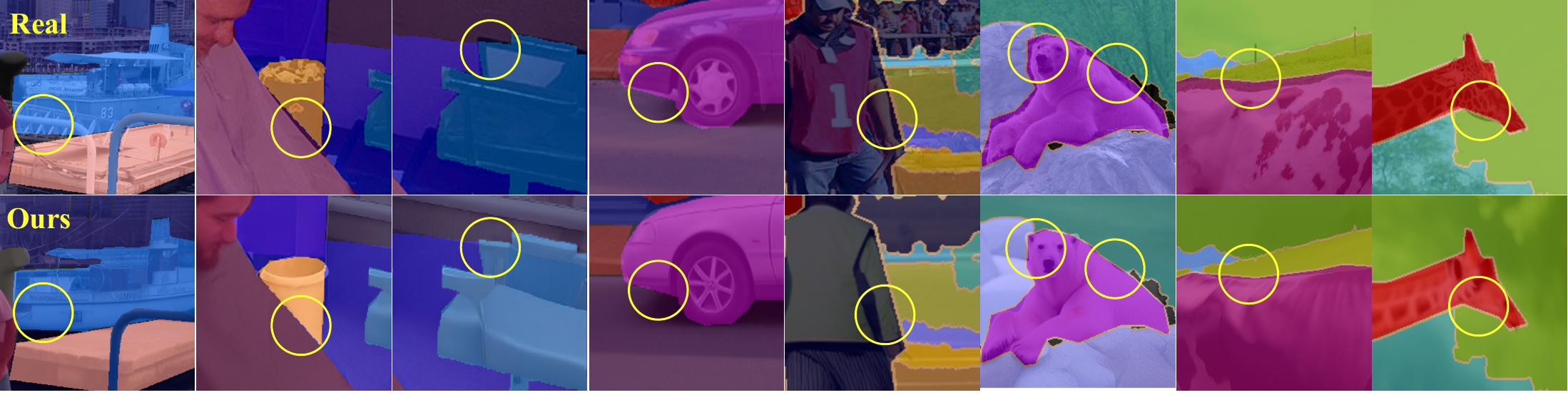}
     \vspace{-20pt}
     \caption{\textbf{Zoom-in comparison of real images and our ImgSyn images:} As highlighted in the circles, our synthetic images align better with the human-labeled masks in many cases because of the inaccuracies in annotations. The left 4 columns are from ADE20K and the right 4 from COCO.}
     \label{fig:better_align}
\vspace{-20pt}
\end{figure*}

\subsection{Data Generation}
With the aforementioned generative models, \modelname proposes two approaches for synthesizing new segmentation training samples: MaskSyn and ImgSyn. 
MaskSyn focuses on enhancing the diversity of synthetic segmentation masks, whereas ImgSyn concentrates on diversifying synthetic images. These generation approaches are illustrated on the right of Fig.~\ref{fig:framework}.

\noindent\textbf{MaskSyn}
MaskSyn starts with a real training sample pair [image, segmentation masks] from a human-annotated segmentation dataset.
It first extracts the caption of the image with the image captioner model.
The caption serves as a text prompt and is used to generate a set of diverse new segmentation masks with the Text2Mask model following Eq.~\ref{eq:txt2mask}. 
Subsequently, the new segmentation masks and the corresponding text prompt are fed into the Mask2Img model to generate a new image that aligns well with its mask.
As such, each training sample crafted by MaskSyn includes both a novel segmentation mask and a new image.
MaskSyn effectively increases the data diversity in segmentation masks to a great extent. Some generated samples are shown in Fig.~\ref{fig:vis_masksyn}.

\noindent\textbf{ImgSyn}
Different from MaskSyn, ImgSyn focuses on increasing the data diversity of images based on human-annotated segmentation masks. For each real training sample pair [image, segmentation mask], it takes the human-annotated segmentation mask and the text prompt extracted from the image as input, and generates a set of varied images that align well with the human-annotated mask.
In this way, new training samples consisting of human-labeled masks and new synthetic images are generated.
ImgSyn can also be viewed as a kind of data augmentation method that enhances the data diversity on the image side.
Our experiments reveal a remarkably high alignment between the synthetic images and their respective segmentation masks. 
Some synthesized results are showcased in Fig.~\ref{fig:vis_imgsyn}.
Our machine-generated synthetic images achieve a better mask-image alignment than real images in many cases, as shown in Fig.~\ref{fig:better_align}.
This phenomenon occurs because human annotations tend to be imperfect due to the high difficulty of annotating segmentation masks. %

\subsection{Training Segmentation Models with Synthetic Data}
The final goal of this work is to improve the performance of current segmentation models with the synthetically generated training data. 
Therefore, we use both synthetic data produced by SegGen and training data from existing datasets in the training process of segmentation models.

We empirically investigate two training strategies utilizing the synthetic data:
{(i) Synthetic data augmentation strategy.}
The synthetic data is used for random data augmentation. 
In every iteration in the training process, each real training sample is replaced by the synthetic training sample with a probability $p_{\textrm{aug}}$.
(ii) Synthetic data pre-training strategy.
It comprises two training stages: pre-training and fine-tuning. In the pre-training stage, we pre-train the segmentation models on synthetic data, so that they learn good weights that are transferable and favorable for fine-tuning.
In the subsequent fine-tuning stage, the segmentation models are trained with solely human-annotated data.

\begin{table}[t]
\vspace{-8pt}
  \centering
  \tablestyle{1.pt}{1.3}
  \scriptsize
  \begin{tabular}{l |cccc|cc}
  \multicolumn{1}{c|}{Method}  & Venue & Backbone & Crop Size & Iterations & mIoU (s.s.) & mIoU (m.s.) \\
  \shline
  MaskFormer~\citep{cheng2021maskformer} & NeurIPS 2021 & R50 & $512$ & 160k & 44.5 & 46.7 \\
  Mask DINO~\citep{li2023maskdino} & CVPR 2023 & R50 & $512$ & 160k & 48.7 & - \\
  
  OneFormer~\citep{jain2023oneformer} & CVPR 2023 & R50 & $512$ & 160k & 47.3 & - \\
  \hline
  Mask2Former~\citep{cheng2021mask2former} & 	CVPR 2022 & R50 & $512$ & 160k & {47.2} & {49.2} \\
  ----w/ \textbf{\modelname (ours)} & - & R50 & $512$ & 160k & {\textbf{49.9} \textbf{\dt{+2.7}}} & {\textbf{51.4 \dt{+2.2}}} \\
  \hline\hline
   MaskFormer~\citep{cheng2021maskformer} & NeurIPS 2021  & Swin-L & $640$ & 160k & 54.1 & 55.6 \\
   Mask DINO~\citep{li2023maskdino} & CVPR 2023 & Swin-L & $640$ & 160k & 56.6 & - \\
   OneFormer~\citep{jain2023oneformer} & CVPR 2023 & Swin-L & $640$ & 160k & 57.0 & 57.7 \\
   \hline
  Mask2Former~\citep{cheng2021mask2former} & 	CVPR 2022  & Swin-L & $640$ & 160k & 56.1 & 57.3 \\
  ----w/ \textbf{\modelname (ours)} & - & Swin-L & $640$ & 160k & {\textbf{57.4} \textbf{\dt{+1.3}}} & {\textbf{58.7 \dt{+1.4}}} \\
  \end{tabular}
  \caption{\textbf{Semantic segmentation on ADE20K \texttt{val} (150 categories):} Synthetic data is used as a data augmentation.
  With the same model structures and training recipes, our \modelname boosts the performance of Mask2Former by a large margin and establishes a new SOTA performance without extra real data under both single-scale~(s.s.) and multi-scale~(m.s.) test settings. }
  \vspace{-24pt}
\label{tab:semseg_ade20k_mask2former}
\end{table}

\section{Experiments}
To accurately evaluate the effectiveness of our data generation method in improving segmentation performance, we adopt mainstream segmentation models and commonly used evaluation benchmarks for several typical segmentation tasks. The experiments are conducted mostly under the fully-supervised learning setting, meaning all human-annotated training samples from the evaluated datasets are used alongside our synthetic data. 
To guarantee an unbiased assessment of the impact of our synthetic data, we keep the architectures of the segmentation models and training protocols consistent with their respective original implementations throughout our studies.

\subsection{Implementation Details}
\label{sec:imple_details}
\textbf{Segmentation Datasets and Evaluation} We conduct experiments on three image segmentation benchmarks following the main experimental settings of Mask2Former~\citep{cheng2021mask2former}: ADE20K semantic segmentation~\citep{ade20k_sceneparse_150},  COCO panoptic segmentation, and COCO instance segmentation~\citep{lin2014coco}.
Our evaluation uses all 150 classes for ADE20K and 133 classes for COCO.
We use all the images from the training splits in the training of segmentation models.
For semantic segmentation, we show the mean Intersection-over-Union metric (mIoU).
For instance segmentation, the average precision (AP) is used.
For panoptic segmentation, we report panoptic quality (PQ), ``thing'' instance segmentation AP$^\text{Th}_\text{pan}$, and semantic segmentation mIoU$_\text{pan}$.

\noindent\textbf{Segmentation Models}
We adopt Mask2Former~\citep{cheng2021mask2former}, a recently-proposed prevalent transformer model, as the default segmentation model.
Two typical backbones, \textit{i.e.}, R50~\citep{he2016deep} and Swin-L~\citep{liu2021swin}, are studied.
We keep the official implementation and training hyper-parameters of the segmentation models unchanged. For more training details please refer to their paper. We also conduct experiments on Mask DINO~\citep{li2023maskdino}, a detection-aided segmentation model, and HRNet W48~\citep{WangSCJDZLMTWLX19hrnet}, a representative fully-convolutional model. More implementation details are introduced in the supplementary materials.

\noindent\noindent\textbf{Data Generation}
The Text2Mask and Mask2Img models are both based on the SDXL-base model~\citep{podell2023sdxl}. They are trained on the training splits of the target segmentation datasets~(\textit{i.e.} ADE20K or COCO) separately for 30,000 iterations using a learning rate of $10^{-5}$. 
AdamW optimizer is employed, and the models are trained at a resolution of 768. Pre-trained weights from SDXL-base~\citep{podell2023sdxl} are utilized. Random flipping data augmentation is used.
During sampling, the default EDM sampler~\citep{karras2022elucidating} is used. The sampling steps are 200 in the Text2Mask model and 40 in the Mask2Img model.
During data sampling, for each training sample in the ADE20K semantic segmentation dataset, we produce 10 synthetic mask-image pairs using MaskSyn, resulting in 202,100 training samples. Additionally, we synthesize 50 images based on each human-labeled mask with ImgSyn, leading to a total of 1,010,500 samples. 
On COCO, we solely use ImgSyn for data synthesis. By generating 10 synthetic images conditioned on each human-labeled panoptic segmentation mask via ImgSyn, our synthetic set amounts to 1,182,870 synthetic samples, which are used in the training of both panoptic and instance segmentation models.

\begin{table}[t]
  \centering
  \resizebox{1.\linewidth}{!}{
  \tablestyle{1pt}{1.2}\scriptsize\begin{tabular}{l | ccc | ccc  cc }
  \multicolumn{1}{c|}{Method}  & Backbone & Queries & Epochs  & PQ & PQ$^\text{Th}$ & PQ$^\text{St}$ & AP$^\text{Th}_\text{pan}$ & mIoU$_\text{pan}$ \\
  \shline
  DETR~\citep{detr} & R50 & 100 & 500+25 & {43.4} & {48.2} & {36.3} & {31.1} & {-} \\
  MaskFormer~\citep{cheng2021maskformer} & R50 & 100 & 300 & 46.5 & 51.0 & 39.8 & 33.0 & 57.8 \\
  Mask2Former~\citep{cheng2021mask2former} & R50 & 100 & 50 & {51.9} & {57.7} & {43.0} & {41.7} & {61.7}  \\
  Mask DINO~\citep{li2023maskdino} & R50 & 300 & 50 & 53.0 & 59.1 & 43.9 & 43.3 & -   \\
  \hline
  Mask2Former~\citep{cheng2021mask2former} & R50 & 100 & 50+50 & 52.0 & 57.9 & 43.4 & 42.0 & 61.0 \\
  --- w/ \textbf{\modelname (ours)}  & R50 & 100 & 50+50 & \textbf{52.7 \dt{+0.7}} & \textbf{58.8 \dt{+0.9}} & \textbf{43.6 \dt{+0.2}} & \textbf{43.1 \dt{+1.1}} & \textbf{62.6 \dt{+1.6}} \\
  \hline
  Mask DINO~\citep{li2023maskdino}  & R50 & 300 & 50+50 & 53.4 & 59.3 & 44.4 & 44.2 & 60.5 \\
  --- w/ \textbf{\modelname (ours)}  & R50 & 300 & 50+50 & \textbf{54.0 \dt{+0.6} }& \textbf{60.2 \dt{+0.9}} & \textbf{44.7 \dt{+0.3}} & \textbf{45.4 \dt{+1.2} }& \textbf{61.5 \dt{+1.0} }\\
  \hline\hline
   MaskFormer~\citep{cheng2021maskformer} & Swin-L & 100 & 300 &  52.7 & 58.5 & 44.0 & 40.1 & 64.8\\
   OneFormer~\citep{jain2023oneformer}  & Swin-L & 150 & 100 & 57.9 & 64.4 & 48.0 & 49.0 & 67.4 \\
  Mask2Former~\citep{cheng2021mask2former} & Swin-L & 200 & 100 & {57.8} & {64.2} & {48.1} & {48.6} & {67.4}  \\
   Mask DINO & Swin-L & 300 & 50 & 58.3 & 65.1 & 48.0 &  50.6 & - \\
   \hline
  Mask2Former~\citep{cheng2021mask2former} & Swin-L & 200 & 100+100 & 57.3 & 64.3 & 46.8 & 48.0 & 66.2 \\
  --- w/ \textbf{\modelname (ours)}  & Swin-L & 200 & 100+100 &  \textbf{58.0 \dt{+0.7}} & \textbf{64.5 \dt{+0.2}} & \textbf{48.1 \dt{+1.3}} & \textbf{48.8 \dt{+0.8}} & \textbf{67.4 \dt{+1.2}} \\
  \hline
  Mask DINO~\citep{li2023maskdino} & Swin-L & 300 & 50+50 & 58.6 & 65.4 & 48.3 & 50.4 & 67.0 \\
  --- w/ \textbf{\modelname (ours)}  & Swin-L & 300 & 50+50 & \textbf{59.3 \dt{+0.7}} & \textbf{65.9 \dt{+0.5}} & \textbf{49.3 \dt{+1.0}} & \textbf{51.1 \dt{+0.7}} & \textbf{68.1 \dt{+1.1}} \\
  \end{tabular}
  }
   \caption{\textbf{Panoptic segmentation on COCO panoptic \texttt{val2017} (133 categories):}  Synthetic data is used for pre-training. Our SegGen significantly improves the performance compared with real data pre-training baselines, and establishes new SOTA performance without extra real data.
\vspace{-10pt}
}

\label{tab:panseg_coco_mask2former}
\end{table}

\subsection{Main Results}

\noindent\textbf{ADE20K Semantic Segmentation}
We employ the synthetic data augmentation strategy with $p_{\textrm{aug}}=60$\% for Mask2Former, and show the results %
in Table~\ref{tab:semseg_ade20k_mask2former}.
SegGen significantly boosts the mIoU of Mask2Former R50 by $\textbf{+2.7}$ for single-scale inference and $\textbf{+2.2}$ for multi-scale inference, achieving 49.9 and 51.4 correspondingly. 
The Swin-L variant is also largely improved from 56.1/57.3~(single-scale/multi-scale) to 57.4/58.7 (\textbf{+1.3/+1.4}).
Notably, our data synthesis method helps Mask2Former surpass the newer methods such as Mask DINO and OneFormer~\citep{jain2023oneformer}, while establishing new SOTA results for R50 and Swin-L settings without using additional human-annotated data.

\noindent\textbf{COCO Panoptic Segmentation}
On COCO we adopt the synthetic data pre-training strategy to utilize our synthetic data. Specifically, we pre-train the models purely on our synthetic data and fine-tune the models on real data.
The results are demonstrated in Table~\ref{tab:panseg_coco_mask2former}.
For a fair comparison, we compare with models pre-trained on purely real data with the same training settings.
Compared with the real data pre-trained baselines, 
SegGen consistently brings performance gains on all metrics with both R50 and Swin-L backbones, achieving new state-of-the-art performance without extra real data. Notably, for Mask DINO with Swin-L backbone, PQ is improved by +0.7, ``thing'' instance segmentation AP$^\text{Th}_\text{pan}$ is increased by +0.7, and semantic segmentation mIoU$_\text{pan}$ is boosted by +1.1. 
These noteworthy findings demonstrate that training with our synthetic data systematically improves performance across various segmentation tasks.
We observe similar significant improvement on Mask2Former, further showing the effectiveness of our synthetic data on different models.

\begin{table*}[t] 
  \centering
\resizebox{1.\linewidth}{!}{
  \tablestyle{1pt}{1.2}\scriptsize\begin{tabular}{l| ccc | x{50}x{50}x{50}x{50} }
  \multicolumn{1}{c|}{Method}  & Backbone & Queries & Epochs & AP & AP$^\text{S}$ & AP$^\text{M}$ & AP$^\text{L}$ \\
  \shline
  Mask R-CNN~\citep{he2017mask} & R50 & anchors & 400 & 42.5 & {23.8} & 45.0 & 60.0 \\
  HTC~\citep{chen2019hybrid}  & R50 & anchors & 36 & 39.7 & 22.6 & 42.2 & 50.6  \\
  QueryInst~\citep{QueryInst}  & R50 & 300 & 36 & 40.6 & 23.4 & 42.5 & 52.8 \\
  MaskFormer~\citep{cheng2021maskformer} & R50 & 100 & 300 & 34.0 & 16.4 & 37.8 & 54.2 \\
  Mask2Former~\citep{cheng2021mask2former}  & R50 & 100 & 50 & {43.7} & 23.4 & {47.2} & {64.8}   \\
  \rowcolor{LightCyan}
  Mask2Former~\citep{cheng2021mask2former}   & R50 & 100 & 50 & 47.1 & 26.6 & 55.4 & 72.7 \\
  \hline
  Mask2Former~\citep{cheng2021mask2former}  & R50 & 100 & 50+50 & 44.1 & 23.4 & 47.7 & {66.1}  \\
  --- w/ \textbf{\modelname  (ours)} & R50 & 100 & 50+50 & \textbf{44.6 \dt{+0.5}} & \textbf{24.0 \dt{+0.6}} & \textbf{48.2 \dt{+0.5}} & \textbf{66.2 \dt{+0.1}} \\
  \rowcolor{LightCyan}
  Mask2Former~\citep{cheng2021mask2former} & R50 & 100 & 50+50 & 47.9 & 27.8 & 56.3 & 73.5  \\
  \rowcolor{LightCyan}
  --- w/ \textbf{\modelname  (ours)} & R50 & 100 & 50+50 & \textbf{48.4 \dt{+0.5}} & \textbf{28.3 \dt{+0.6}} & \textbf{57.4 \dt{+1.1}} & \textbf{74.1 \dt{+0.6}}  \\
  \hline\hline
  Swin-HTC++~\citep{liu2021swin} & Swin-L & anchors & 72 &  49.5 & {31.0} & 52.4 & 67.2 \\
  QueryInst~\citep{QueryInst} & Swin-L & 300 & 50 &  48.9 & 30.8 & 52.6 & 68.3  \\
  Oneformer~\citep{jain2023oneformer}  & Swin-L & 150 & 100 & 48.9 & - &  - & - \\
  Mask2Former~\citep{cheng2021mask2former}  & Swin-L & {200} & {100} & {50.1} & 29.9 & {53.9} & {72.1}  \\
  \rowcolor{LightCyan} 
  Mask2Former~\citep{cheng2021mask2former}  & Swin-L & {200} & {100} & 54.3 & 35.7 & 63.3 & 79.1  \\
  \hline
  Mask2Former~\citep{cheng2021mask2former}  & Swin-L & {200} & 100+100 & 49.5 & 29.2 & 53.8 & 70.5 \\
  --- w/ \textbf{\modelname (ours)}  & Swin-L & 200 & {100+100} & \textbf{50.3 \dt{+0.8}}& \textbf{31.2 \dt{+2.0}} & \textbf{54.3 \dt{+0.5}} & \textbf{72.2 \dt{+1.7}} \\
  \rowcolor{LightCyan}
   Mask2Former~\citep{cheng2021mask2former}  & Swin-L & {200} & 100+100 & 53.6 & 35.2 & 62.6 & 77.8 \\
  \rowcolor{LightCyan}
   --- w/ \textbf{\modelname (ours)}  & Swin-L & 200 & {100+100} & \textbf{54.8 \dt{+1.2}} & \textbf{35.9 \dt{+0.8}} & \textbf{64.2 \dt{+1.6}} & \textbf{79.2 \dt{+1.5}}  \\
  \end{tabular}
  }
   \caption{\textbf{Instance segmentation on COCO \texttt{val2017} (80 categories):}
   Synthetic data is used for pre-training.
   The gray rows correspond to the evaluations conducted on LVIS~\citep{gupta2019lvis} which offers instance masks of higher quality.
   Our SegGen significantly improves the performance of instance segmentation for various sizes of objects compared with pre-training on real data.
   }
\vspace{-30pt}
\label{tab:insseg_coco_mask2former}
\end{table*}

\begin{figure}[!t]
\begin{minipage}[t]{0.28\textwidth} %
    \includegraphics[width=\linewidth]{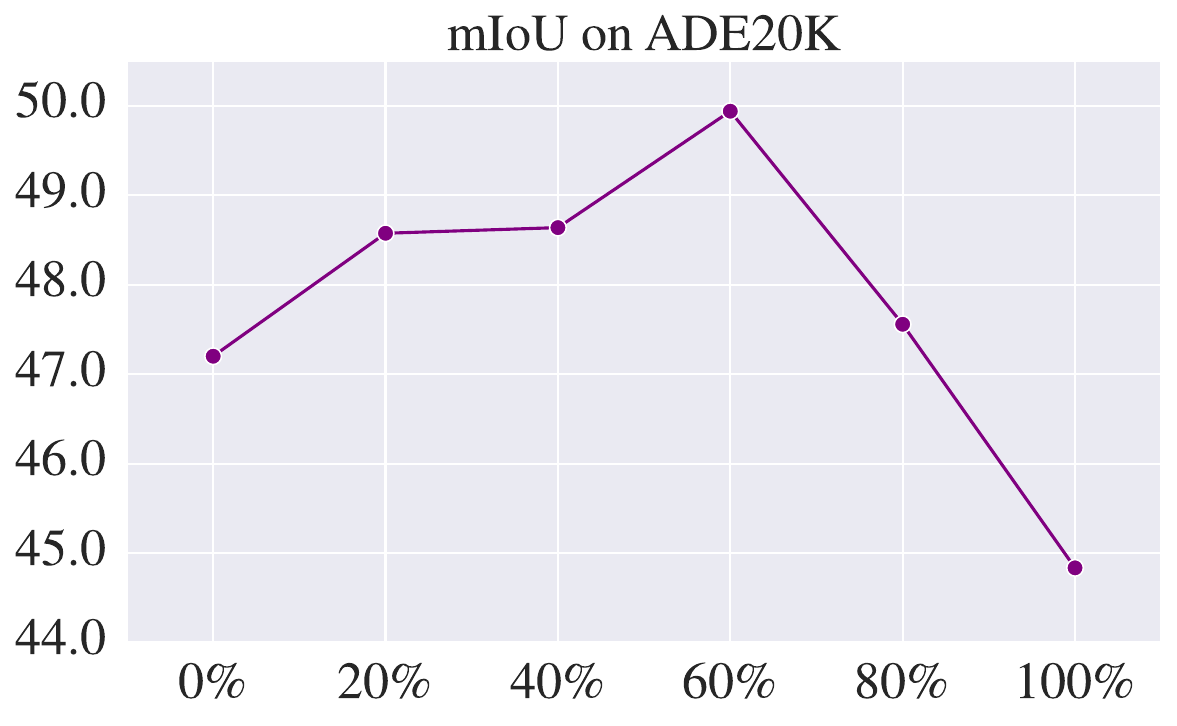}
    \vspace{-20pt}
    \captionof{figure}{ADE20K mIoU of Mask2Former R50 with different augmentation probability $p_{\textrm{aug}}$.}
    \label{fig:ablation_aug_prob}
\end{minipage}%
\hfill
\begin{minipage}[t]{0.28\textwidth} %
    \includegraphics[width=\linewidth]{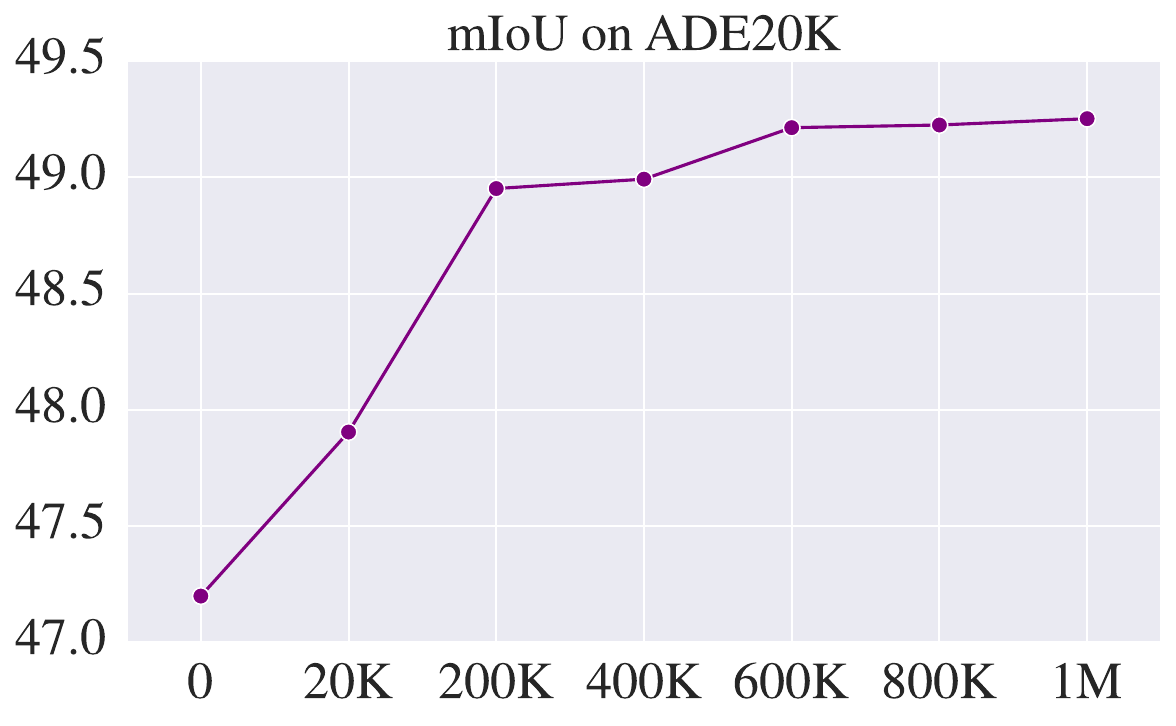}
    \vspace{-20pt}
    \captionof{figure}{ADE20K mIoU of Mask2Former R50 using different numbers of synthetic samples.}
    \label{fig:ablation_imgsyn_size}
\end{minipage}%
\hfill
\begin{minipage}[t]{0.4\linewidth} %
    \centering
    \vspace{-60pt}
    \tablestyle{1pt}{1.2}
    \scriptsize
    \resizebox{1.\linewidth}{!}{
    \begin{tabular}{l| x{50}x{20}x{20}|x{20}}
          \multicolumn{1}{c|}{Method} & mIoU* \\
          \shline
          Pure Real Data & 83.3 \\
          \hline
          DiffuMask (ICCV 2023)  & 57.0 \\
          MaskSyn+ImgSyn (ours) & \textbf{82.2\dt{+25.2}}\\
          MaskSyn (ours)  & 73.2 \\
          ImgSyn (ours)  & \underline{80.0} \\
        \end{tabular}
        }
        \vspace{-10pt}
        \captionof{table}{\textbf{Comparison with DiffuMask~\citep{wu2023diffumask} on ADE20K:} All models trained purely on \textit{synthetic data}. mIoU* is defined in~\citep{wu2023diffumask}. }
        \label{tab:peer_compare}
\end{minipage}
\vspace{-20pt}
\end{figure}

\noindent\textbf{COCO Instance Segmentation}
The performance on instance segmentation, both before and after incorporating our SegGen synthetic data, is detailed in Table~\ref{tab:insseg_coco_mask2former}.
Compared to the baselines pre-trained on real data, our technique is superior across all metrics. 
Specifically, there is a solid +0.5 improvement in AP with the R50 backbone, and a +0.8 increase when using the Swin-L backbone.
Additional assessment on the LVIS~\cite{gupta2019lvis} annotated COCO validation set, whose annotation is more accurate than the original version, provides additional evidence of the effectiveness of SegGen, as it enhances the average precision (AP) by +1.2 when used with the larger Swin-L backbone.

\noindent\textbf{Comparison with Segmentation Data Generation Method}
In Table~\ref{tab:peer_compare}, we compare various versions of our SegGen with DiffuMask~\citep{wu2023diffumask}, which is a recently-proposed (ICCV2023) segmentation data generation approach.  All methods employ the Mask2Former R50 model and are \textit{purely trained on synthetic samples}. We adhere to the evaluation setting designed by DiffuMask where the IoU for three common classes are examined. Our method markedly surpasses DiffuMask by \textbf{+25.2 mIoU} and demonstrates performance comparable to training purely with real data, highlighting the unprecedented quality of our synthetic data.

\begin{figure*}[t]
    \centering
     \includegraphics[width=1\linewidth]{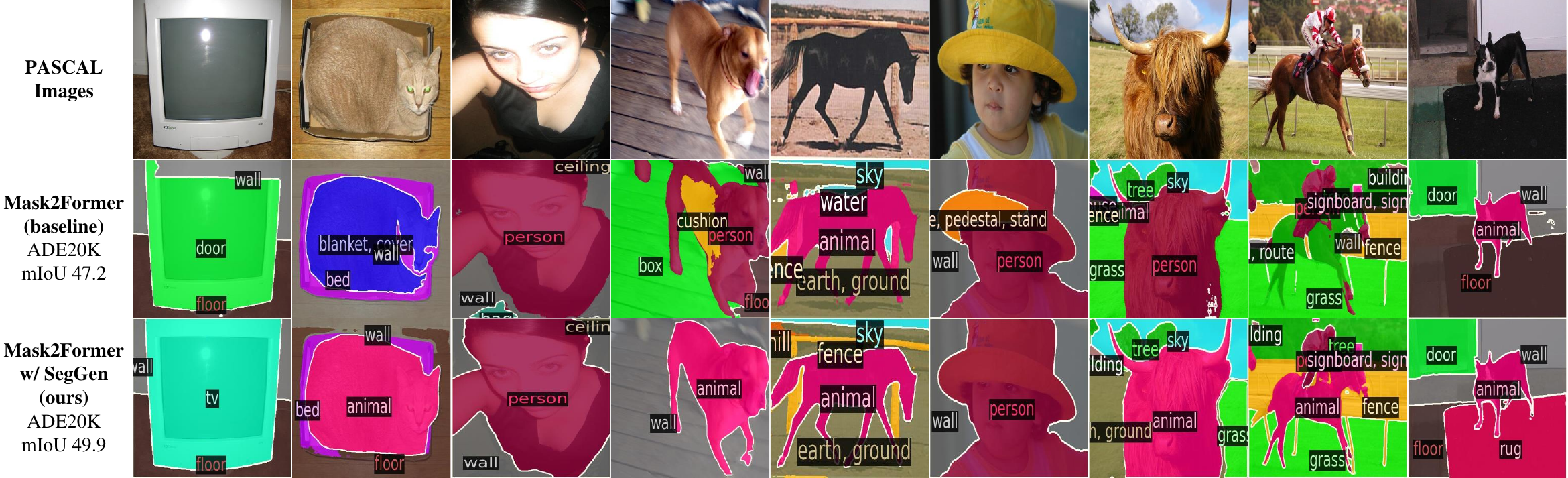}
     \vspace{-20pt}
     \caption{\textbf{Generalization Ability on unseen domains}: Segmentation outputs on unseen domain~(images from PASCAL~\citep{everingham2015pascal}). The models are trained on ADE20K. Training with our \modelname demonstrates enhanced robustness towards unfamiliar domains.}
     \label{fig:compare_pascal}
\vspace{-11pt}
\end{figure*}

\noindent\textbf{Generalization Ability on Unseen Domains}
We visualize the segmentation results of Mask2Former R50 on images from PASCAL \texttt{val} dataset~\citep{everingham2015pascal} in Fig.~\ref{fig:teaser} and Fig.~\ref{fig:compare_pascal}, comparing models trained both with and without our synthetic data.
The models are trained on COCO or ADE20K.
When trained using our extensively varied synthetic data, the segmentation model demonstrates notably improved performance on the unseen domain. %
We further study the segmentation results on AI-generated images, which are synthesized using different image generation models, in the supplementary materials.

\noindent\textbf{Visual Analysis of Generated Samples by MaskSyn}
We randomly select some samples generated by MaskSyn in Fig.~\ref{fig:vis_masksyn} (more in supplementary materials).
The generative models are trained on ADE20K.
Given the text prompts, our Text2Mask model can generate diverse segmentation masks, and the Mask2Img model produces realistic synthetic images with a good alignment with the masks and text prompts.

\noindent\textbf{Visual Analysis of Generated Samples by ImgSyn}
We showcase the outputs by ImgSyn in Fig.~\ref{fig:vis_imgsyn} (more in supplementary materials).
The generated images are in good agreement with the text prompts and human-labeled segmentation masks.
Furthermore, we compare the mask-image alignment of the real training samples and our synthetic training samples in Fig.~\ref{fig:better_align}. Our synthetic samples exhibit superior alignment quality in many cases, compared to human annotations which are often imperfect on the boundaries.

\begin{table}[t]
  \begin{subtable}{0.49\linewidth}
    \centering
    \tablestyle{1pt}{1.2}
    \scriptsize
    \resizebox{1.\linewidth}{!}{
    \begin{tabular}{l|cc|c}
          \multicolumn{1}{c|}{Method} & Iterations & Real Samples & mIoU \\
          \shline
          Baseline  & 160K & 20210 & 47.2\\
          --- w/ \textbf{\modelname} MaskSyn  & 160K & 20210 & 48.5 \\
          --- w/ \textbf{\modelname} ImgSyn  & 160K & 20210 & \underline{49.3} \\
          --- w/ \textbf{\modelname} ImgSyn + MaskSyn  & 160K & 20210 & \textbf{49.9} \\
          \hline
          Baseline & 160K & 1000 & 23.7 \\
          --- w/ \textbf{\modelname} MaskSyn  & 160K & 1000 &  \underline{26.5}  \\
          --- w/ \textbf{\modelname} ImgSyn  & 160K & 1000 & 24.1 \\
          --- w/ \textbf{\modelname} ImgSyn + MaskSyn  & 160K & 1000 & \textbf{27.1}  \\
        \end{tabular}
        }
        \vspace{-4pt}
    \caption{Mask2Former R50}
    \end{subtable}
    \hfill
    \begin{subtable}{0.49\linewidth}
    \centering
    \tablestyle{1pt}{1.2}
    \scriptsize
    \resizebox{1.\linewidth}{!}{
    \begin{tabular}{l|cc|c}
              \multicolumn{1}{c|}{Method} & Iterations & Real Samples & mIoU \\
              \shline
              Baseline & 100K & 20210 & 44.4 \\
              --- w/ \textbf{\modelname} MaskSyn & 100K  & 20210 &  44.8  \\
              --- w/ \textbf{\modelname} ImgSyn & 100K  & 20210 & \underline{45.4} \\
              --- w/ \textbf{\modelname} ImgSyn + MaskSyn & 100K  & 20210 & \textbf{45.6} \\
              \hline
              Baseline & 100K & 1000 & 23.6 \\
              --- w/ \textbf{\modelname} MaskSyn & 100K  & 1000 &  \underline{26.0}  \\
              --- w/ \textbf{\modelname} ImgSyn & 100K  & 1000 & 23.9 \\
              --- w/ \textbf{\modelname} ImgSyn + MaskSyn & 100K  & 1000 & \textbf{26.9} \\
        \end{tabular}
        }
    \vspace{-4pt}
    \caption{HRNet W48}
    \end{subtable}
    \vspace{-8pt}
    \caption{\textbf{Ablation Study of MaskSyn and ImgSyn on ADE20K:} Both MaskSyn and ImgSyn notably enhance the performance of Mask2Former and HRNet, while MaskSyn has a more pronounced impact when there are fewer real samples available.}
    \label{tab:ablation_seggen}
    \vspace{-28pt}
\end{table}

\subsection{Quantitative Analysis}

\textbf{Effectiveness of MaskSyn and ImgSyn}
We evaluate the data generated by MaskSyn and ImgSyn separately and jointly, by examining their impacts on the performance of segmentation models, as shown in Table~\ref{tab:ablation_seggen}. The experiments are conducted on ADE20K with Mask2Former R50 and HRNet W48.
We find that the training samples generated by both MaskSyn and ImgSyn bring significant performance gains compared with the baseline. When combining these two types of synthetic data together, the segmentation model can achieve the best performance.
We delve deeper into a scenario where only 1,000 real training samples are available in the training of both generative models and segmentation models. 
It is observed that MaskSyn substantially enhances the segmentation results in situations with limited real data. This could be attributed to the ability of MaskSyn to amplify data diversity by creating entirely new segmentation masks and images.
Moreover, the significant performance improvements achieved by our SegGen, trained on merely 1,000 real samples, demonstrate the data efficiency and robustness of our proposal.

\noindent\textbf{Influence of Synthetic Data Augmentation Probability \(p_{\text{aug}}\) on ADE20K}
We investigate the effect of \(p_{\text{aug}}\) in Fig.~\ref{fig:ablation_aug_prob}. When \(p_{\text{aug}} = 0\), it implies training using only real data, whereas \(p_{\text{aug}} = 100\%\) means training entirely with synthetic data. The best results are achieved at \(p_{\text{aug}} = 60\%\). It is noteworthy that training exclusively on our synthetic data results in an impressive 44.8 mIoU.

\noindent\textbf{Influence of Synthetic Data Size}
To evaluate the impact of synthetic data volume, we use different numbers of synthetic training samples produced by ImgSyn in training and report the results in Fig.~\ref{fig:ablation_imgsyn_size}. $p_{\textrm{aug}}$ is fixed at 60\%.
We notice a substantial improvement in performance when increasing the synthetic data quantity from 20K to 200K, underscoring the importance of collecting a significantly larger synthetic dataset to avoid overfitting. The performance appears to plateau after 600K. More studies on the scale of synthetic data are presented in the supplementary materials. %

\noindent\textbf{Influence of Training Strategies}
We conduct an ablation study using different training strategies with synthetic data on ADE20K~(Table~\ref{tab:ablation_train_strategy_ade20k}) and COCO~(Table~\ref{tab:ablation_train_strategy_coco}).
We find that using the synthetic data augmentation strategy achieves better performance on ADE20K semantic segmentation,
While on COCO panoptic and instance segmentation, the synthetic pre-training strategy works better. 
We believe that due to the limited size of the ADE20K training set, which contains merely $\sim$20k images, segmentation models easily overfit the scarcely-available training images and segmentation layouts. %
Utilizing synthetic data~(including synthetic masks and images) for data augmentation can help mitigate such an overfitting problem. On the other hand, for COCO, given its larger training set size (around 100K images), overfitting is less of a concern. 
Hence, it is sufficient to provide the COCO models with good initial weights pre-trained on our synthetic data.
Meanwhile, previous work~\citep{gupta2019lvis} finds that the annotation accuracy of COCO is significantly more biased than ADE20K, a finding that aligns with our visualization results depicted in Fig.~\ref{fig:better_align}. A domain gap exists between the COCO data and our high-quality synthetic data. Hence, employing our method for pre-training and incorporating real data for fine-tuning on COCO is a more favorable approach.

\begin{table}[t]
    \begin{subtable}{0.49\linewidth}
    \centering
    \tablestyle{1.pt}{1.2}
    \scriptsize
        \begin{tabular}{p{2.cm}|x{60}|x{30}}
          \multicolumn{1}{c|}{Method} & Iterations & mIoU \\
          \shline
          Syn. Pre-Train & 160K+160K & 47.2 \\
          Syn. Data Aug & 160K & \textbf{49.9} \\
        \end{tabular}
        \caption{\textbf{ADE20K Semantic Segmentation:} Using synthetic data as random data augmentation obtains the best performance on ADE20K.
        \phantom{pad}\phantom{pad}\phantom{pad}
        }
        \label{tab:ablation_train_strategy_ade20k}
    \end{subtable}
    \hfill
    \begin{subtable}{0.49\linewidth}
    \centering
    \tablestyle{1.pt}{1.2}
    \scriptsize
        \begin{tabular}{p{2.5cm}|x{40}| x{20}x{20}}
          \multicolumn{1}{c|}{Method} & Epochs & PQ & AP \\
          \shline
          Syn. Pre-Training & 50+50 & \textbf{52.7} & \textbf{44.6} \\
          Syn. Data Aug & 50 & 51.3 & 43.2 \\
        \end{tabular}
          \caption{\textbf{COCO Panoptic Segmentation and Instance Segmentation:} Using synthetic data for pre-training achieves better performance on COCO dataset.
         } %
        \label{tab:ablation_train_strategy_coco}
    \end{subtable}
    \vspace{-8pt}
    \caption{Different training strategies with synthetic data using Mask2Former R50.}
    \vspace{-28pt}
\end{table}

\section{Conclusion}
We present SegGen, a highly-effective data synthesis method for image segmentation. 
SegGen introduces a revolutionary segmentation data generation framework, and proposes two data generation approaches, namely MaskSyn and ImgSyn, to synthesize segmentation masks and images, with the help of the designed text-to-mask and mask-to-image generation models.
Our method not only strongly enhances the model performance on semantic, panoptic, and instance segmentation benchmarks, but also substantially improves segmentation generalization ability in unseen data distribution.

\clearpage  %

\bibliographystyle{splncs04}
\bibliography{main}

\clearpage

\appendix
\begin{center}
    \LARGE\textbf{Appendix}
\end{center}
\addcontentsline{toc}{section}{Appendix}

\section{Generalization Ability on AI-Generated Images}
\begin{figure*}[ht]
    \centering
    \vspace{-10pt}
     \includegraphics[width=1.\linewidth]{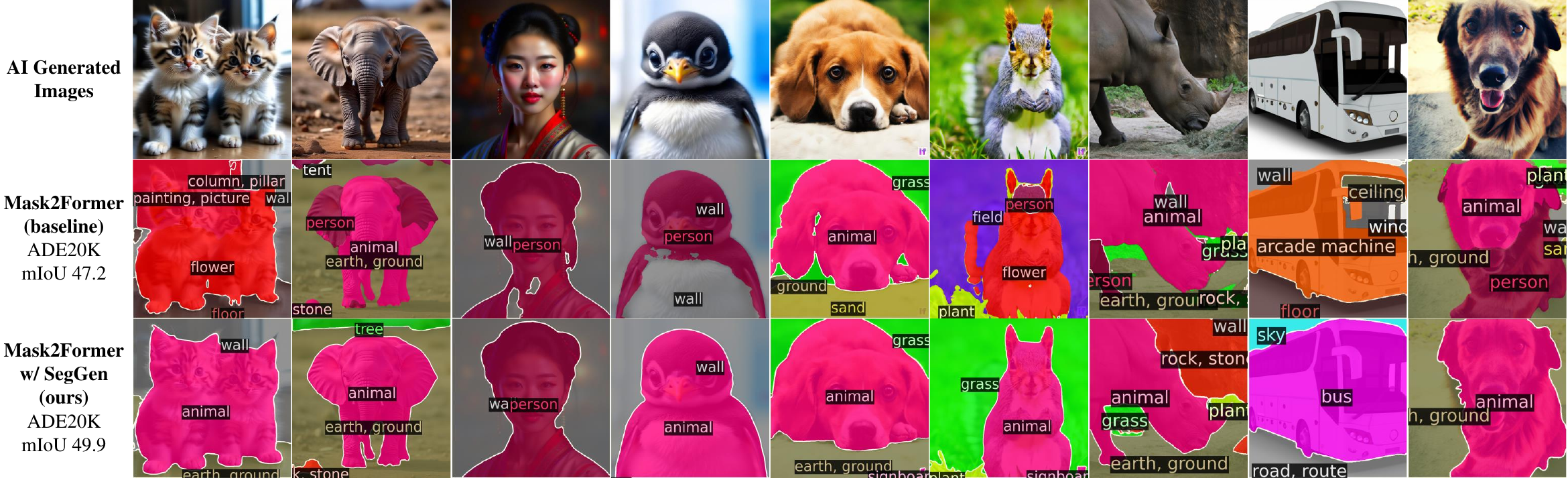}
     \vspace{-15pt}
     \caption{\textbf{Generalization ability on AI-generated images}: Segmentation outputs of Mask2Former model~\cite{cheng2021mask2former} before and after training with our SegGen on ADE20K. The first 3 columns of images are generated by Kandinsky 2, the middle 3 by IF, and the later 3 by DALL-E. Training with our synthetic data brings significantly stronger generalization ability to the segmentation model.}
     \label{fig:compare_gen_img}
\vspace{-20pt}
\end{figure*}

We randomly generate a set of high-quality images with the latest text-to-image generation methods: 
Kandinsky 2 by \cite{kandinsky}, IF by \cite{if}, and DALL-E by \cite{dalle2021}.
We visualize the segmentation results of Mask2Former R50 trained both with and without our synthetic data in Fig.~\ref{fig:compare_gen_img}.
The models are trained on ADE20K.
When trained using our comprehensive synthetic data, the segmentation model obtains notably improved performance on the images synthesized by generative models, showing its stronger generalization ability.

\section{Visual Analysis of Segmentation Color-Map Reconstruction Capability of VAE}
\begin{figure*}[h]
    \centering
    \begin{subfigure}[b]{0.5\linewidth}
        \centering
        \includegraphics[width=1.\linewidth]{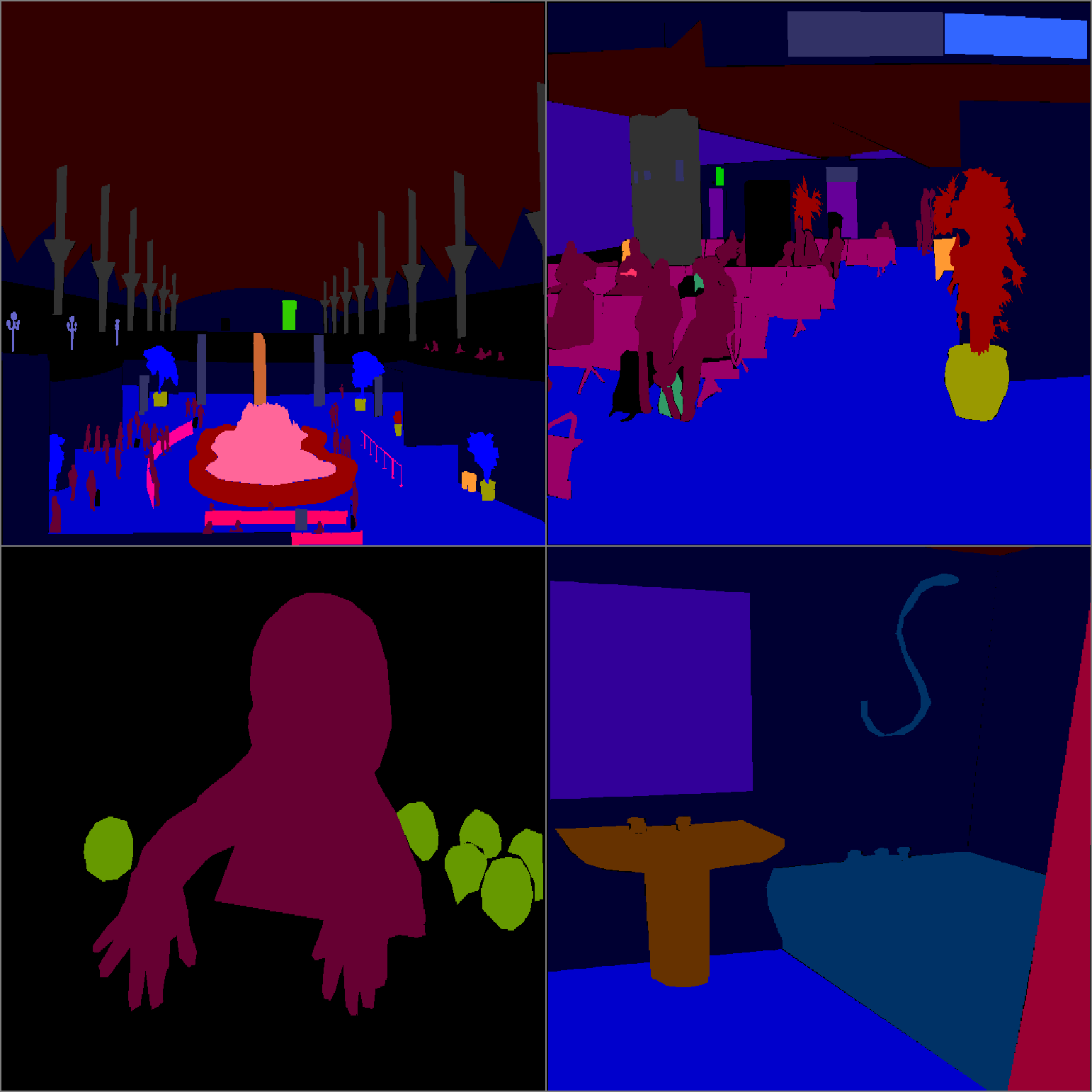}
        \caption{VAE Inputs}
    \end{subfigure}%
    ~ %
    \begin{subfigure}[b]{0.5\linewidth}
        \centering
        \includegraphics[width=1.\linewidth]{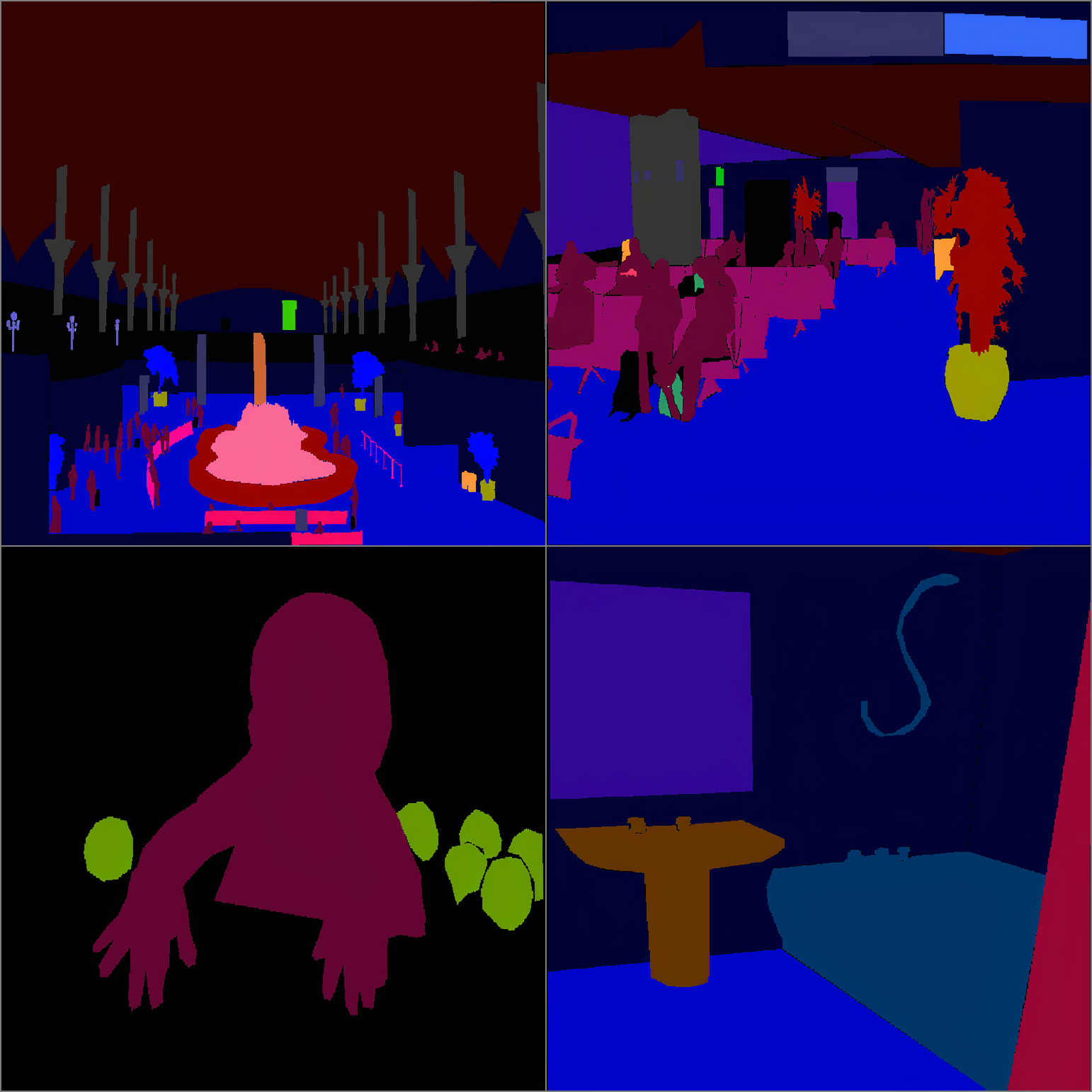}
        \caption{VAE Reconstructions} %
    \end{subfigure}
     \vspace{-15pt}
    \caption{\textbf{Visual analysis of segmentation color map reconstruction using VAE:} The pre-trained VAE of SDXL can effectively reconstruct color maps, establishing the foundation of our Text2Mask model.}
    \label{fig:vae}
\end{figure*}

We show the inputs and reconstruction outcomes of variational autoencoders~(VAE) in our Text2Mask model in Fig.~\ref{fig:vae}. %
The VAE, which is borrowed from SDXL~\cite{podell2023sdxl}, demonstrates an impressive ability to reconstruct the segmentation color maps, enabling our training of the Text2Mask generation model.

\section{More Implementation Details}

\subsection{More Implementation Details about Segmentation Models}
\label{sec:imple_seg}

\noindent\textbf{Mask2Former} We follow the official training scripts and settings from their GitHub repository. 
The initial learning rate is $10^{-4}$, weight decay rate is $5\times 10^{-2}$, and the batch size is 16. For more implementation details please refer to \cite{cheng2021mask2former}. %

\noindent\textbf{Mask DINO} Mask DINO is a recently proposed segmentation transformer aided by object detection networks. We use the official training scripts available on their GitHub repository and adhere to their training configurations. The starting learning rate is set at $10^{-4}$, the weight decay rate at $5\times 10^{-2}$, and a batch size of 16. For specifics on the loss structure and further implementation details, kindly consult \cite{li2023maskdino}.

\noindent\textbf{HRNet} For experiments with HRNet~\cite{WangSCJDZLMTWLX19hrnet}, we use the W48 version, and train it for 100K iterations. SGD optimizer is used with a starting learning rate of  $2\times 10^{-2}$. A polynomial learning rate scheduler is used with a power parameter set to 0.9. The weight decay rate is $10^{-4}$. Random flipping data augmentation is incorporated.

\subsection{More Implementation Details about Generative Models}
\label{sec:appendix_imple_gen}
The Text2Mask and Mask2Img models are both based on the SDXL-base model~\cite{podell2023sdxl}. They are trained on the training splits of the target segmentation datasets~(\textit{i.e.} ADE20K or COCO) separately for 30,000 iterations using a learning rate of $10^{-5}$. 
AdamW optimizer is employed, and the models are trained at a resolution of 768$\times$768. Pre-trained weights from SDXL-base~\cite{podell2023sdxl} are utilized. Random flipping data augmentation is used.
During sampling, the default EDM sampler~\cite{karras2022elucidating} is used. The number of sampling steps is 200 in the Text2Mask model and 40 in the Mask2Img model.

\section{More Study on Synthetic Data Scale}
\label{sec:more_study_synthetic_size}
\begin{figure}[h]
    \centering
    \vspace{-10pt}
    \begin{subfigure}[b]{0.48\textwidth}
        \centering
        \includegraphics[width=1\linewidth]{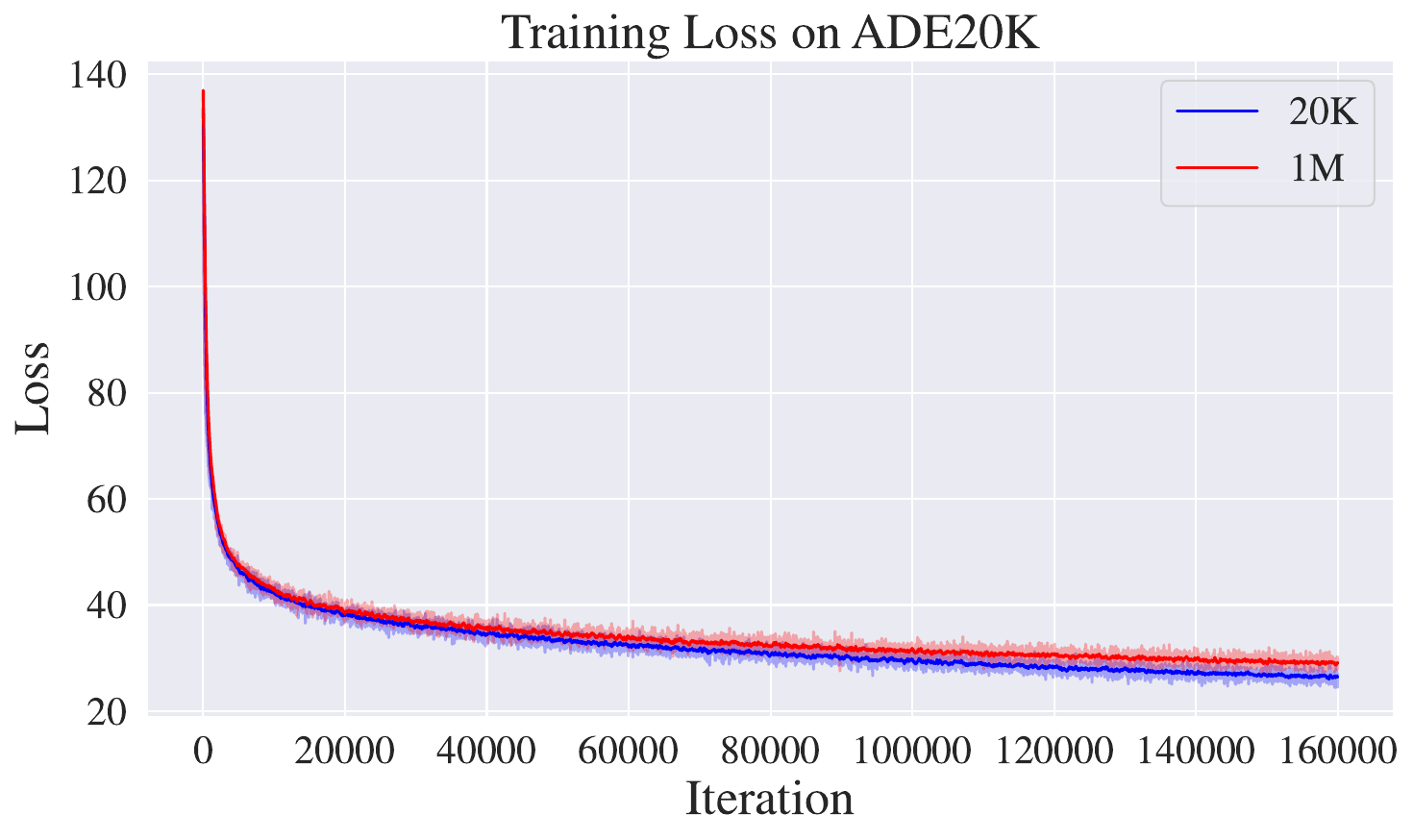}
        \caption{Training Loss}
        \label{fig:sub1}
    \end{subfigure}
    \hfill %
    \begin{subfigure}[b]{0.48\textwidth}
        \centering
        \includegraphics[width=1\linewidth]{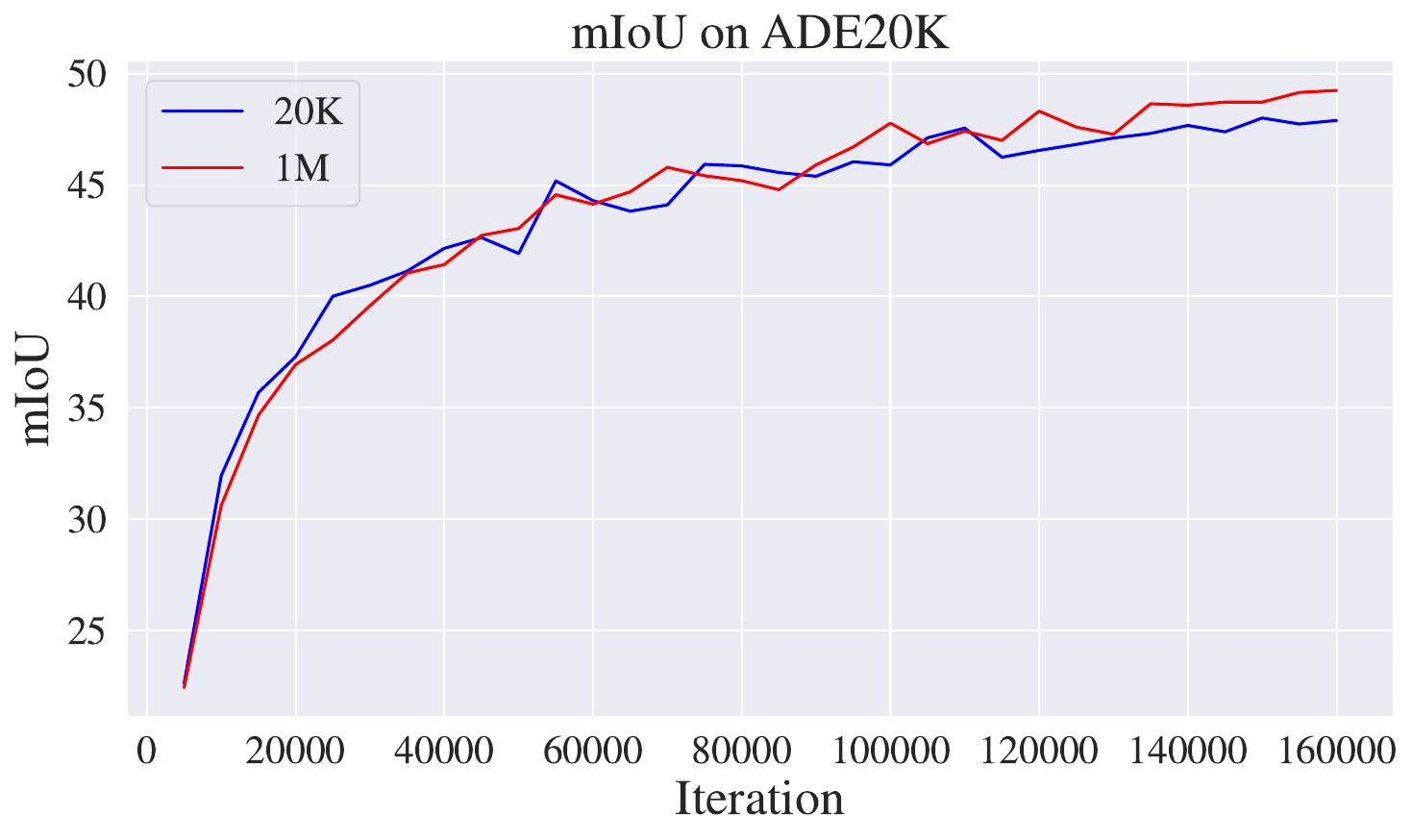}
        \caption{Validation mIoU}
        \label{fig:sub2}
    \end{subfigure}
    
    \caption{\textbf{Training loss and validation mIoU on ADE20K:} Using a limited amount of synthetic data (\textit{e.g.} 20K) can result in overfitting and subsequently worse performance. Therefore we generate more than 1M synthetic training samples, which significantly mitigates the issue. This experiment highlights the importance of the scale of synthetic data.} 
    \label{fig:syn_data_size_curve}
    \vspace{-10pt}
\end{figure}

To better understand the impact of synthetic data scale, we train the Mask2-Former R50 model using different quantities of ImgSyn samples~(\textit{i.e.} 20K and 1M), and show the training loss and validation mIoU in Fig.~\ref{fig:syn_data_size_curve}. The synthetic data augmentation probability remains constant at 60\%.

We find that utilizing 20K synthetic samples results in a lower training loss compared to using 1M samples; however, the validation mIoU is significantly better with the larger synthetic dataset. 
\textbf{The cause behind this phenomenon is:}
Statistically, when we use only 20K synthetic samples, the model encounters each synthetic sample approximately 76.8 times during training. In contrast, with 1M synthetic samples, each sample is presented to the model about 1.5 times. This indicates that when the synthetic dataset is not large enough, there is a risk of the model overfitting to the synthetic data.
Hence, employing a more extensive synthetic dataset is very beneficial for training segmentation models, as it prevents the model from easily memorizing all the synthetic training samples.
Our SegGen can easily scale up the synthetic data by sampling new masks and images using the proposed generative models. This is an insignificant cost compared to hiring human annotators to annotate segmentation data.

\section{Effectiveness of SegGen without Data Augmentation}
In order to verify the efficacy of our approach without employing conventional data augmentation techniques, we carry out an experiment where we eliminate color augmentation, random flipping, and random cropping from Mask2Former R50. We train it using our SegGen data on ADE20K. The outcomes of this experiment are presented in Table~\ref{tab:wo_data_aug}.
Our observations demonstrate that in the absence of data augmentation methods, our synthetic data has a more pronounced impact on enhancing the segmentation performance. Notably, it brings a notable increase in mIoU by +4.8.

\begin{table}[t]
    \centering
    \tablestyle{8pt}{1.3}
    \scriptsize
    \begin{tabular}{l|cc|c}
          \multicolumn{1}{c|}{Method} & Iterations  & Data Aug  & mIoU \\
          \shline
          Baseline  & 160K & Yes & 47.2\\
          --- w/ \textbf{\modelname} & 160K & Yes & \textbf{49.9 (+2.7)} \\
          \hline
          Baseline  & 160K & No & 41.4 \\
          --- w/ \textbf{\modelname} & 160K & No & \textbf{46.2 (+4.8)} \\
        \end{tabular}
    \caption{\textbf{Effectiveness of SegGen without data augmentation:} When these data augmentation methods are unavailable, our SegGen brings a more significant improvement.}
    \label{tab:wo_data_aug}
\end{table}

\section{Impact of Different Amounts of MaskSyn Samples}
We conduct a series of experiments to examine the impact of using different amounts of MaskSyn samples, as shown in Table~\ref{tab:ablation_masksyn_size}. We use Mask2Former R50 as the segmentation model. Similar to the observations in experiments for ImgSyn in the main text, the performance is consistently improved with the increase of MaskSyn samples. This further underscores the importance of generating large-scale segmentation data to train more effective segmentation models.

\begin{table}[h]
    \centering
    \tablestyle{8pt}{1.3}
    \scriptsize
    \begin{tabular}{l|cc|c}
          \multicolumn{1}{c|}{Method} & Iterations  & MaskSyn Samples  & mIoU \\
          \shline
          Baseline  & 160K & 0 & 47.2\\
          \textbf{\modelname} MaskSyn & 160K & 40K  & 47.3 \\
          \textbf{\modelname} MaskSyn & 160K & 80K  & 47.6 \\
          \textbf{\modelname} MaskSyn & 160K & 160K  & 48.1 \\
          \textbf{\modelname} MaskSyn & 160K & 200K  & 48.5 \\
        \end{tabular}
    \caption{\textbf{Influence of different numbers of MaskSyn samples in training Mask2Former R50:} More synthetic MaskSyn samples leads to notably better performance.}
    \label{tab:ablation_masksyn_size}
\end{table}

\section{Visual Analysis of Generated Samples}

\begin{figure*}[!t]
    \centering
     \includegraphics[width=1.\linewidth]{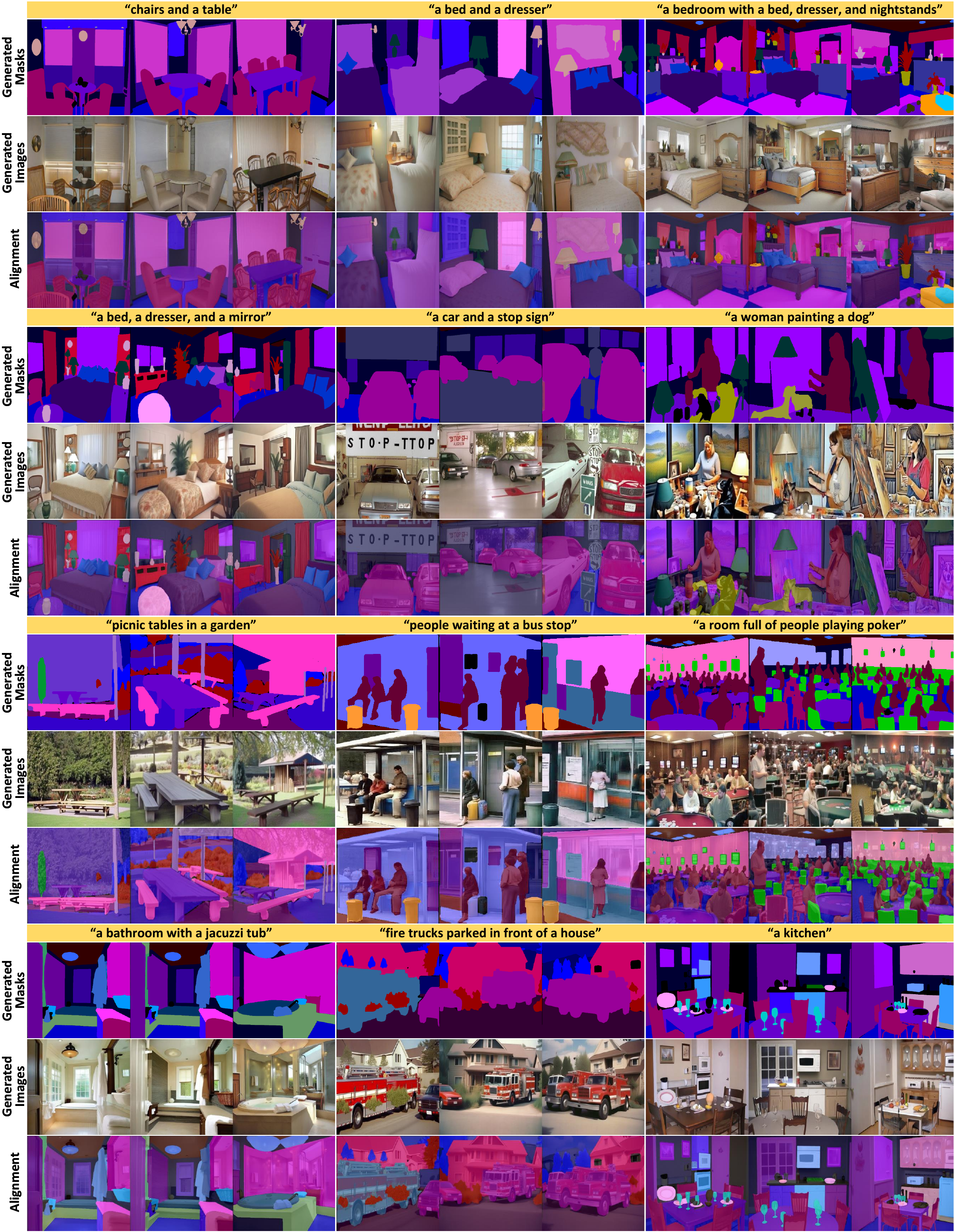}
     \vspace{-15pt}     
     \caption{\textbf{More generated samples by MaskSyn on ADE20K:} Our synthetic segmentation masks are highly diverse, and the synthetic images align well with the respective masks and text prompts.}
     \label{fig:vis_masksyn_more}
\end{figure*}

\begin{figure*}[!t]
    \centering
     \includegraphics[width=1.\linewidth]{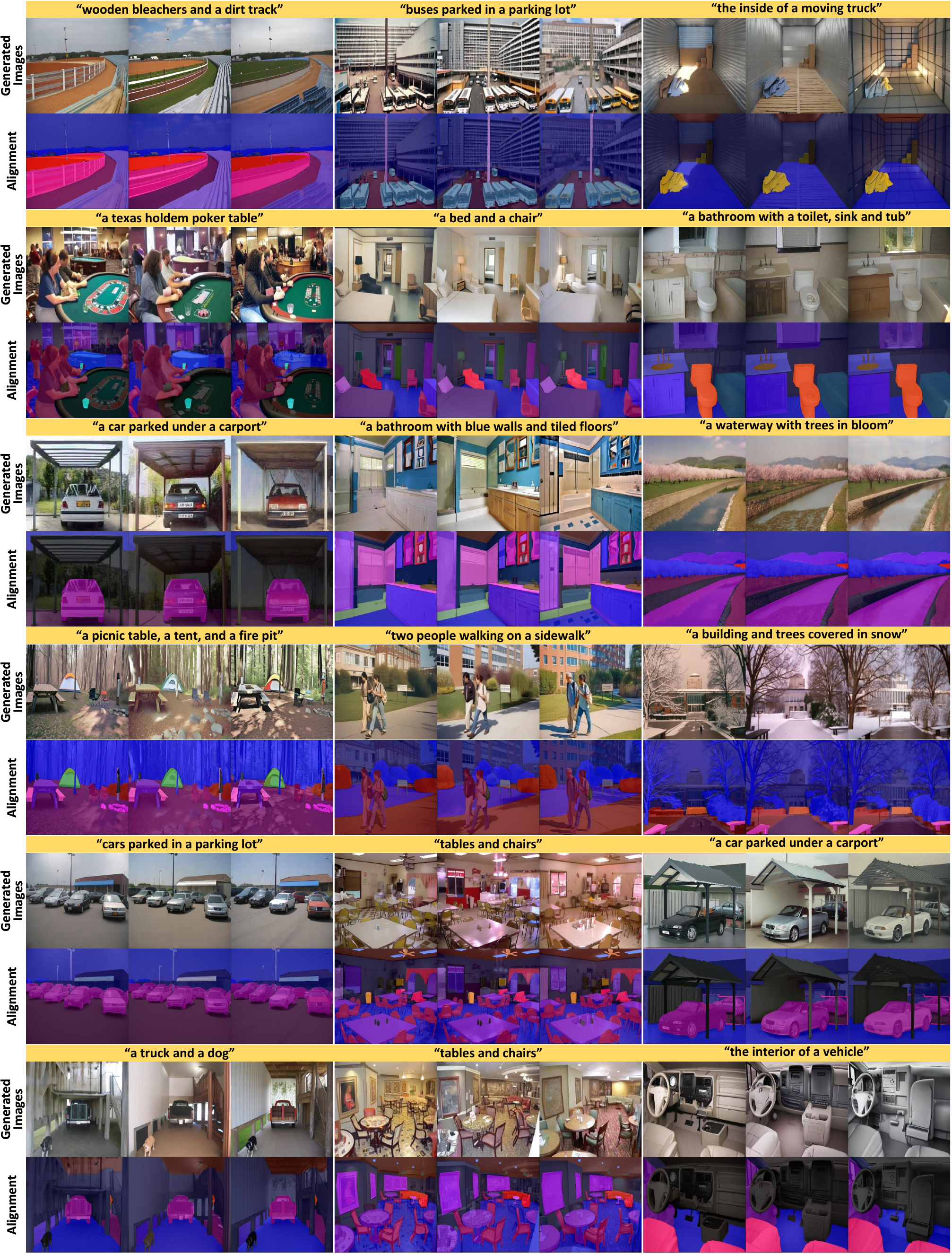}
     \vspace{-15pt}
     \caption{\textbf{More generated samples by ImgSyn on ADE20K:} The generated images align well with the text prompts and human-annotated segmentation masks.}
     \label{fig:vis_imgsyn_more}
\end{figure*}

\begin{figure*}[!t]
    \centering
     \includegraphics[width=1.\linewidth]{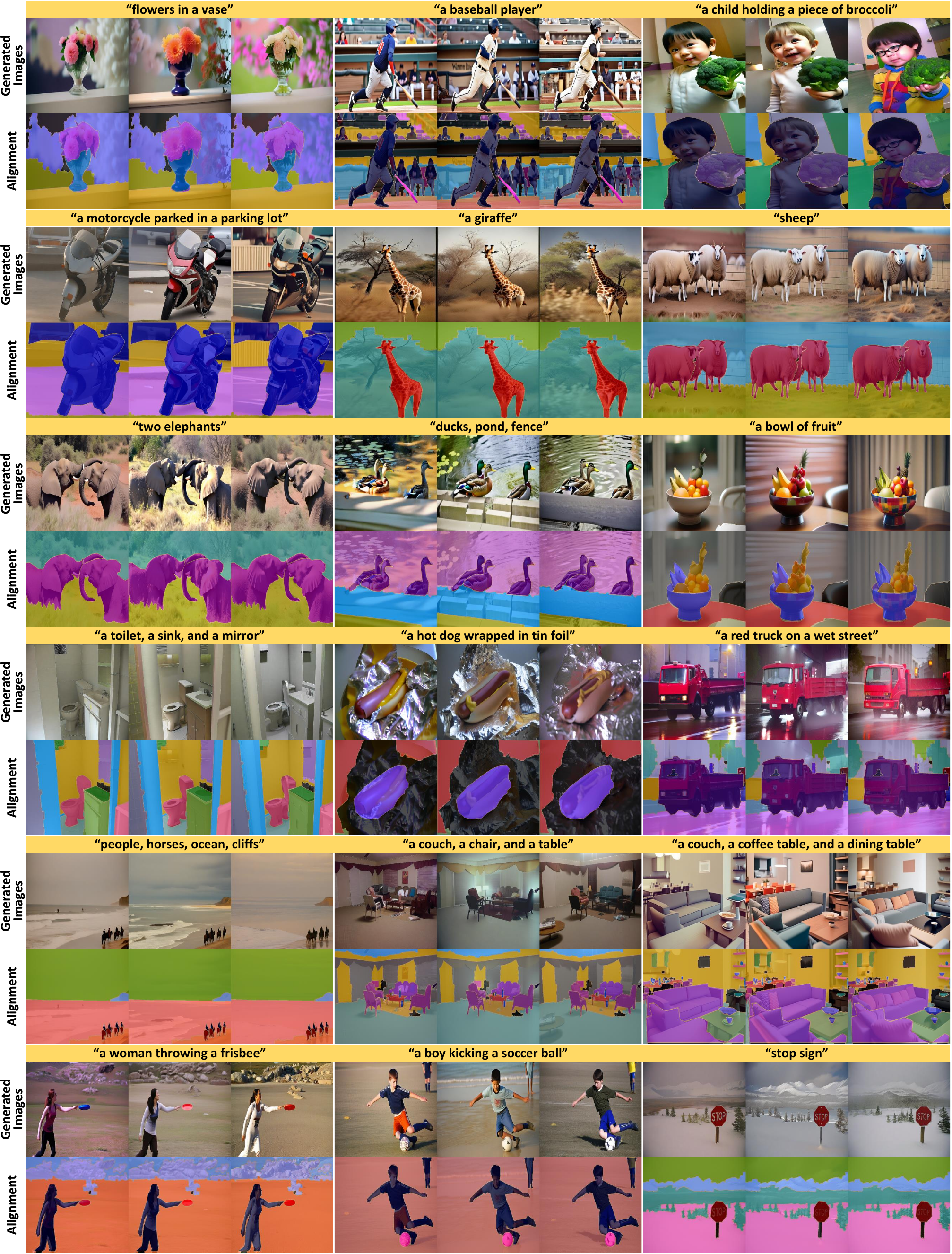}
     \vspace{-20pt}
     \caption{\textbf{More generated samples by ImgSyn on COCO:} The generated images align well with the text prompts and human-annotated segmentation masks.}
     \label{fig:vis_imgsyn_coco}
\end{figure*}

In Fig.\ref{fig:vis_masksyn_more} and Fig.\ref{fig:vis_imgsyn_more}, we present additional samples generated by our MaskSyn and ImgSyn on the ADE20K dataset. Fig.~\ref{fig:vis_imgsyn_coco} displays the generated samples on the COCO dataset. These illustrations highlight the superior quality of the synthetic masks and images produced by our \modelname.

\end{document}